\documentclass[journal]{IEEEtran}

\usepackage{amsmath, amsfonts}
\usepackage{algorithm}
\usepackage{algorithmic}
\usepackage{array}
\usepackage[caption=false, font=normalsize, labelfont=sf, textfont=sf]{subfig}
\usepackage{textcomp}
\usepackage{stfloats}
\usepackage{graphicx}
\usepackage{cite}

\usepackage{amssymb}
\usepackage{bm} % for bold symbols
\usepackage{multirow}
\usepackage{hyperref}
\hypersetup{hidelinks}
\usepackage{siunitx}
\usepackage[dvipsnames]{xcolor}
\usepackage{balance}
\usepackage{tikz}

\newtheorem{remark}{Remark}

\newcommand{\vect}[1]{\ensuremath{\bm{#1}}}
\newcolumntype{?}[1]{!{\vrule width #1}}\def \q {{\vect{q}}}
\def \F {{\vect{F}}}
\def \z {{\vect{z}}}
\def \u {{\vect{u}}}
\def \w{{\vect{w}}}
\def \y {{\vect{y}}}
\def \x {{\vect{x}}}
\def \d {{\vect{d}}}
\def \n {{\vect{n}}}
\def \r {{\vect{r}}}
\def \e {{\vect{e}}}

\begin{document}

\title{\LARGE \bf 
Optimal Virtual Model Control for Robotics: \\ 
Design and Tuning of Passivity-Based Controllers
}

\author{Daniel Larby,~\IEEEmembership{Member,~IEEE,} Fulvio Forni,~\IEEEmembership{Senior Member,~IEEE,}%
    \thanks{Manuscript received XXXX XX, XXXX; revised XXXXX XX, XXXX. This work was supported by the Engineering and Physical Sciences Research Council  [EP/T517847/1]; and by CMR Surgical. For the purpose of open access, the author has applied a Creative Commons Attribution (CC BY) licence to any Author Accepted Manuscript version arising.}
    \thanks{Fulvio Forni is with the Department of Engineering, University of Cambridge, CB2 1PZ, UK. Daniel Larby was also the Department of Engineering, University of Cambridge, but is now with Swan Endosurgical, 155 Cambridge Science Park, Cambridge CB4 0GN. (e-mail: dan\_larby@hotmail.co.uk; f.forni@eng.cam.ac.uk)}
}

\markboth{IEEE TRANSACTIONS ON ROBOTICS, VOL. X, NO. X, XXXXXXX 202X}%
{Shell \MakeLowercase{\textit{et al.}}: A Sample Article Using IEEEtran.cls for IEEE Journals}

% \IEEEpubid{0000--0000/00\$00.00~\copyright~2025 IEEE}

\maketitle
\begin{abstract}
Passivity-based control is a cornerstone of control theory and an established design approach in robotics. Its strength is based on the passivity theorem, which provides a powerful interconnection framework for robotics. However, the design of passivity-based controllers and their optimal tuning remain challenging. We propose here an intuitive design approach for fully actuated robots, where the control action is determined by a `virtual-mechanism' as in classical virtual model control. The result is a robot whose controlled behavior can be understood in terms of physics. We achieve optimal tuning by applying algorithmic differentiation to ODE simulations of the rigid body dynamics. Overall, this leads to a flexible design and optimization approach: stability is proven by passivity of the virtual mechanism, while performance is obtained by optimization using algorithmic differentiation. 
\end{abstract}

\section{Introduction}

\IEEEPARstart{P}{assivity}-based control methods are crucial for designing of stable and reliable robot controllers
\cite{Takegaki1981, Ortega2001, Ortega2008, Chopra2022, Secchi2007, Slotine1987, Ortega1989}. 
Passivity-based approaches consider the energy of the robot-controller interconnection (energy shaping) and regulate its dissipation (damping injection) to ensure stability. The aim is a controlled robot  that is passive, thus stable when interacting with other passive devices. The passivity theorem \cite{vanderSchaft2017, Sepulchre1997} guarantees stability even if both the robot model and the environment are uncertain, provided they are passive.

A classical example is proportional derivative (PD) control plus gravity compensation in end-effector space \cite{Khatib1987,Ortega1998}. The proportional action has the mechanical interpretation of a spring. Combined with gravity compensation, it shapes the potential energy of the controlled robot. The derivative action has the mechanical interpretation of a damper, which has the effect of slowing down the robot motion towards the minimum of the shaped potential energy (stabilization via damping injection). The resulting controlled robot is passive and very reliable. In fact, small perturbations or uncertainties in the robot dynamics would not affect the robot motion in a significant way. 

Although passivity-based control is a well-established control approach, in applications few practical designs go beyond PD control and quadratic potentials. In fact, related (and powerful!) approaches  such as virtual fixtures \cite{Bowyer2014} and impedance controllers too often remain within the comfort of proportional-derivative reaction to the position error. In general, it is difficult to connect the design of a complex controller with the specific features of the task. In control-theoretic terms, the challenge is finding the relevant task-dependent metric and then optimizing the structure and parameters of the controller for this metric. Picture the design of a controller for a surgical laparoscopic task \cite{Su2019, Marinho2019, Su2020a}, where the surgical tool is constrained to pass through a small, but somewhat compliant aperture into the patient's body. Precise position or force control are insufficient to tackle the complex interaction required by the task. Attacking this through the mathematical derivation of a specific potential energy of passivity-based control appears to fall short. In fact, what kind of `energy-shaping' is best for this task? Even when the controlled impedance of the robot is considered \cite{Hogan1984}, the desired impedance is often chosen heuristically \cite{Hogan2022}. Pragmatically, the review \cite{Song2019} concludes that efforts are needed ``(...) to establish more systematic guidance or methodology on how to specify impedance parameters [such as desired inertia, damping, and stiffness] to reflect the basic dynamic interaction requirements of robotic manipulations raised from customers''.

\IEEEpubidadjcol % This is here in the second column of the first page to make space for the IEEE publisher ID at the bottom of the page

Our hypothesis is that passivity-based control remains a central design approach in these complex settings. The question is how to explore the space of passivity-based controllers in terms of task-oriented design and optimal tuning. 
We propose a two-stage approach: first, design the controller as a \emph{virtual mechanism} connected to the robot; second,
tune it using \emph{scientific machine-learning} by differentiation of the rigid-body-dynamics through simulations. 

The design of controllers as virtual mechanisms is the philosophy of virtual model control \cite{Pratt1995b, Joly1995}. 
Conceptualizing the controller as a virtual mechanism falls within the framework of control through interconnection \cite{Ortega2008, Bloch2000, Ortega1998, Ortega2004}, but opens the design space to mechanical intuition and the ability to customize the structure of the controller to the specific features of the task.
Examples of virtual model control have been developed in locomotion of bipedal walking robots \cite{Pratt1998, JianjuenHrr1998},
exoskeletons \cite{Ekkelenkamp2007}, quadrupeds \cite{Desai2014, Chen2020a}, virtual fixture approaches \cite{Bowyer2014}
and telerobotics \cite{Joly1995, Xu2016}.
One of the contributions of this paper is to revisit virtual model control design at fundamental level, starting from the systematic characterization of its basic elements, to arrive to the task-oriented design of passive controllers. 

\begin{figure}[t]
    \centering
    \includegraphics[width=\columnwidth]{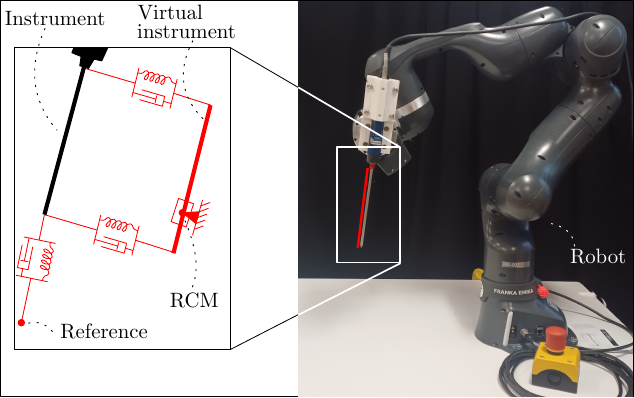}
    \caption{The virtual mechanism for laparoscopic surgery. 
    The motion of the virtual `instrument', in red, is constrained to a pre-defined 
    remote centre of motion (RCM) through  prismatic and rotatory joints. The virtual instrument is `attached' to the robot through springs and dampers. Springs are shown in a stretched configuration, at equilibrium there would be no distance between the virtual instrument and the surgical tool.}
    \label{fig:VirtualInstrument}
\end{figure}

For the case of laparoscopic surgery, we can design a compliant virtual mechanism that connects to the surgical tool, constraining its interaction with soft tissues. An illustration is provided in Fig. \ref{fig:VirtualInstrument}.
This is neither position nor force control. It shows connections with impedance control, but without enforcing a specific desired impedance. The shaping of the energy is not reduced to the manipulation of mathematical terms, but is determined by the geometry and dynamics of the virtual mechanism and by the virtual interconnection with the surgical tool, both driven by mechanical intuition.
For the specific case of laparoscopic surgery, our approach shows connections with energy shaping RCM control approaches \cite{Su2020a, Su2020, Su2019, Sadeghian2019, Sandoval2018a, aghakhani_task_2013, Funda1996}. However, our energy-shaping is driven (and constrained) by the geometrical and mechanical representation of a virtual mechanism. This is a qualitative difference which leads to the ability to parametrize entire families of suitable controllers, that can be optimised in a task-oriented way.

Often in robotic control, nonlinearities are seen as problematic dynamics to be removed.
In contrast, in a virtual mechanisms controller, a nonlinear spring can stiffen or saturate at different extensions, therefore providing complex controlled feedback forces to serve different performance requirements.
Nonlinear mechanical linkages become a tool for shaping the kinetic and potential energy of the system intuitively without designing custom distance functions for energy-shaping potential fields.
Virtual mechanisms are easy to implement using virtual operation space forces. Their implementation is easy to debug, as the controller action can be understood in terms of the physics of the virtual mechanism.
If the virtual mechanism is passive, then the closed loop is passive, guaranteeing convergence to a minima of the potential energy of the controlled robot. In fact, the virtual mechanism approach provides robustness for the sim-to-real transition: even with geometric/inertial modeling errors, so long as the robot and controller are passive, we have some level of stability guarantee. 

For optimal tuning, we are inspired by recent work on scientific-machine learning: the practise of combining scientific models with gradient-based optimization \cite{Rackauckas2021}.
Recent developments in automatic differentiation of ordinary differential equation (ODE) solvers have opened the door to efficient optimizations involving continuous-time dynamical systems, attracting increasing attention in robotics \cite{Rackauckas2021a, Kim2021, Chen2019,Howell2023,Giftthaler2017, Giftthaler2018}. 
Scientific machine learning has been used in related applications in physical system identification \cite{Raissi2019, Greydanus2019, Brunton2019};
to find optimal state feedback control policies \cite{Sandoval2022};
to perform optimal energy shaping for a pendulum, tuning a controller which is a passive neural network \cite{Massaroli2022}; to design an energy shaping controller to generate desired closed-loop oscillatory trajectories on an undamped double pendulum \cite{Wotte2022}. 

In this paper, we embrace this philosophy. 
We consider the robot as an \emph{open system} whose
input/output behavior must be optimized. 
The optimization of the controller does not rely on general-purpose architectures, such as neural networks, autoencoders, etc. In contrast, our `model' (in the sense of machine learning) is the virtual mechanism designed by the user. 
The expressivity of the controller, that is, the ability to generate a wide family of control policies, is regulated by the level of parameterization of the virtual mechanism, ranging from stiffness and damping parameters to length, inertia, and other geometrical features of its mechanical linkages.
These parameters are optimized using loss functions that are surrogates of
classical performance control metrics, namely the $\mathcal{L}_2$ and $\mathcal{L}_\infty$ metrics \cite{Zhou1998}.
Compared to \cite{Sandoval2022},
the use of virtual mechanisms leads to dynamic feedback policies. Passivity
is not enforced through constraints as in \cite{Massaroli2022}, but inherited from the structure of the controller. 

Computing a gradient through an ODE solver `properly' allows us to compute quickly,
thus handle the optimization over the dynamics of the controlled robot efficiently.
Using this approach to tune virtual mechanisms can meet the expectations of the review \cite{Song2019} for systematic parameter specification.
At the same time, our approach guarantees stability/robustness certificates, which are often missing in state-of-the-art machine learning methods (sim-to-real gap), limiting their applicability in critical domains such as surgical robotics, where a higher level of certainty about controller stability/robustness is required.

The contributions of this paper include:
(i) Design of passivity-based controllers as virtual mechanisms, discussed in Sections \ref{sec:VirtualMechanisms} and \ref{sec:VirtualMechInControl}. These sections extend the early results of \cite{Larby2022}.
(ii) The adaptation of $\mathcal{L}_2$ and $\mathcal{L}_\infty$ performance metrics for tractable optimization based on automatic differentiation. This is discussed in Section \ref{sec:CostFunctions}.
(iii) A tuning approach based on scientific machine learning implemented by algorithmic differentiation of continuous-time simulations, discussed in Section
\ref{sec:Tuning}. This extends the approach of \cite{Larby2022b}, based on linear matrix inequalities \cite{Boyd1994}.
(iv) Simulation results for the 1DOF and 7DOF robot in a surgical setting.
We also provide experimental validation, as shown in Section \ref{sec:SurgeryExample}.

\section{Passivity Fundamentals}
Even if passivity is widely used and well known in the context robot control \cite{Takegaki1981, Ortega2001, Ortega2008, Chopra2022, Secchi2007, Slotine1987, Ortega1989,vanderSchaft2017, Sepulchre1997},  we briefly revisit the main concepts
to make this paper self-contained.
Mechanical systems without internal sources are passive, that is, \emph{any variation of their mechanical energy is bounded by the energy supplied externally}.
Power may flow into/out of a mechanical system through interconnection ports consisting of a co-located (generalized) force and velocity $(\bm{F}, \dot{\bm{z}})$.
A mechanical system without internal sources and with a single port $(\bm{F}, \dot{\bm{z}})$ satisfies.
\begin{equation}
    E(t_1) - E(t_0)\le  \int_{t_0}^{t_1} \dot{\bm{z}}^T \bm{F} dt,
\end{equation}
where $E$ represents the mechanical energy of the system, and $t_0 \leq t_1$ are generic time instants. This inequality is the footprint of a passive system. It shows how the variation of mechanical energy is bounded by the supplied energy. Without external supply, the energy of the mechanical system cannot increase (no internal sources). 
When the energy of the system is differentiable, this is equivalent to 
\begin{equation}
    \label{eq:PassivityDt}
    \dot{E} \le \dot{\bm{z}}^T  \bm{F}. 
\end{equation}

The interconnection of passive mechanical system is also passive. To see this, consider two passive mechanical systems with energy $E_1$ and $E_2$, respectively. The interconnection of these system through two generic ports, $(\bm{F}_1, \dot{\bm{z}_1})$ on the first system and $(\bm{F}_2, \dot{\bm{z}_2})$ on the second system, must 
satisfy either, 
\begin{subequations}
\begin{align}
\dot{\bm{z}}_1 = \dot{\bm{z}}_2 = \dot{\bm{z}}
\qquad 
\bm{F}_1 + \bm{F}_2 = \bm{F},
\end{align}
or 
\begin{align}
\dot{\bm{z}}_1 + \dot{\bm{z}}_2 = \dot{\bm{z}}
\qquad
\bm{F}_1 = \bm{F}_2 = \bm{F},
\end{align}
\end{subequations}
where $(\bm{F},\bm{z})$ are additional co-located force and velocity variables.
Then, the interconnected system has total energy
\begin{equation}
E = E_1 + E_2
\end{equation}
and is passive with respect to the `new' port $(\bm{F}, \dot{\bm{z}})$.
For instance, 
\begin{equation}
\dot{E} = \dot{E}_1 + \dot{E}_2
\le \dot{\bm{z}}_1^T \bm{F}_1 + \dot{\bm{z}}_2^T \bm{F}_2 
\le \dot{\bm{z}}^T \bm{F}
\end{equation}

The feature that passivity is preserved by mechanical interconnections is at the core of the use of virtual mechanisms for control design. The idea is that robots are passive and can be controlled by `interconnecting' them to virtual mechanisms, which are also passive. Later, the action of this virtual mechanism is realized through the controlled action of the available motor actuation. 

It is a standard result of passive system theory that, in the absence of external forces and in the presence of sufficient damping, all trajectories of the controlled robot converge to some resting state. These correspond to the equilibria of the system, given by zero velocity and by generalized positions that are extremal of the potential energy of the controlled system. 
This is because without power flowing into the system, $\bm{F}=0 \implies \dot{\bm{z}}^T \bm{F} = 0$, the rate of change of energy is strictly negative\footnote{Due to the `sufficient' damping mentioned above} unless the system is at rest \cite{Ortega1989, Ortega1998}.

\section{Virtual mechanisms fundamentals} 
\label{sec:VirtualMechanisms}

\subsection{Virtual forces}
\label{sec:VirtualForces}
To design a controller as a virtual mechanism we need a way to realize the
virtual forces arising from the interconnection of virtual mechanism and robot manipulator. 
These virtual forces must be emulated through the available actuation. 
Thus, for fully actuated robots, the objective is to drive the actuators of the robot to realise the effect that these virtual forces \emph{would} cause on the dynamics of the robot, using standard methods from Cartesian and operation space control \cite{Khatib1987}, \cite{Spong1989}, \cite{Siciliano1999}. In what follows, we confine the discussion to virtual forces for reasons of simplicity, but the approach can be extended to include torques.

\begin{figure}[t]
    \centering
    \includegraphics[page=2]{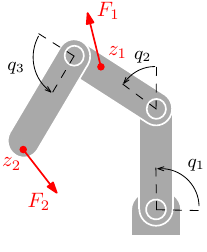}
    \caption{Virtual forces $\F_1$ and $\F_2$ acting on points $\z_1$ and $\z_2$.}
    \label{fig:OperationSpaceControl}
\end{figure}

Consider the robot in Fig. \ref{fig:OperationSpaceControl} with generalized coordinates  $\q$.
The generic point $\z$ is a point rigidly attached to one of the robot's links.
We want to realize the effect of a virtual force $\F$ at $\z$.
The position of $\z$ in some robot frame (e.g. the robot's base frame) can be represented as a function of the generalized coordinates $\q$, namely $\z = h_{\z}(\q)$.
The velocity reads 
\begin{equation}
    \label{eq:AbstractCoordinateVelocity}
    \dot{\z} = J_{\z}(\q) \dot \q
\end{equation}
where $J_{\z}(\q) = \frac{\partial h_{\z}(\q)}{\partial \q}.$ 
Both $\F$ and $\z$ are vectors of three elements.
By the principle of virtual work \cite{Spong1989}, the joint actuation 
\begin{equation}
\label{eq:AbstractCoordinateTorque}
\u = J_{\z}(\q)^T \F
\end{equation}
makes the robot move as if driven by the force $\F$ acting at point $\z$, where $\F$ is represented in the robot frame of $\dot{\z}$.

The power flow through $(\u, \dot \q)$ and $(\F, \dot \z)$ is conserved
\begin{align}
    \F^T \dot \z = \F^T J_\z(\q) \dot \q = \u^T \dot \q.
\end{align}
The pairs $(\F, \dot \z)$ and $(\u,\dot{\q})$ are ports \cite{vanderSchaft2020} and the Jacobian $J_\z(\q)$ models a lossless interconnection between them. 
The virtual force $\F$, and its effect on the robot dynamics, are then physically realized via $\u$. 

The effect of multiple virtual forces $\F_i$ at generic points $\z_i$, $i\in\{1,\dots,n\}$, is captured by superposition\cite{Hogan1984}:
\begin{equation}
    \u = \sum_{i=1}^n J_{\z_i}(\q)^T \F_i.
\end{equation}

% As  shown in Fig. \ref{fig:OperationSpaceControl}, in what follows 
% virtual forces will be generated by virtual components 
% connected between two points, $\z_1$ and $\z_2$, of the robot.
% In this case, considering the incremental coordinate $\z_3 = \z_1 - \z_2$, 
% we have
% \begin{alignat}{3}
%     & \z_3       &&=  h_{\z_1}(\q) - h_{\z_2}(\q) \\
%     & \dot \z_3  &&= (J_{z_1}(\q)- J_{\z_2}(q)) \dot \q \\
%     & \u         &&= (J_{z_1}(\q)- J_{\z_2}(\q))^T \F_3,
% \end{alignat}
% where $\F_3$ is the force in the diin opposite directions\footnote{Note that in Fig. \ref{fig:OperationSpaceControl}, $\F_3$ is shown as acting parallel to $z_3$, which is not always the case, but is typical if the component inducing $\F_3$ is a spring or damper.} on both $\z_2$ and $z_1$.

% \begin{remark}
% In full generality, we may define a port for any coordinate for which for which we can define a velocity as \eqref{eq:AbstractCoordinateVelocity}.
% The corresponding virtual force $F$, which is in the same space as $\dot z$, can then be applied using \eqref{eq:AbstractCoordinateTorque}.
% This is sufficient for power flow through $(\u, \dot \q)$, $(\F, \dot \z)$to be preserved.
% A function $h(q)$ to compute configuration $z$ is not strictly necessary.
% \end{remark}

% \begin{remark}
% If a coordinate is defined so that $\dot \z=\vect{\omega}$, the angular velocity of a link, then the corresponding $\F$ represents a torque applied to the link, rather than a linear force.
% \end{remark}

\subsection{Virtual elements}

\begin{figure}[t]
    \centering
    \includegraphics[]{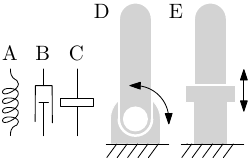}
    \caption{Components symbols. A) Spring. B) Damper. C) Inerter. D) Revolute joint. E) Prismatic joint.}
    \label{fig:components}
\end{figure}

The virtual mechanism is built of primitive mechanical elements such as ideal springs, dampers, and inerters combined into virtual structures.
Fig. \ref{fig:components} shows the symbols adopted in this paper. For simplicity we restrict the discussion to one degree-of-freedom joints, but the approach can be easily extended to more general joints. 

\subsubsection*{Springs}
An ideal spring is defined by its potential energy function $V:\mathbb{R}^3 \to \mathbb{R}$, which must be lower bounded for the spring to be passive. 
Let the displacement $\z = \z_1 - \z_2$ be the extension of a virtual spring
whose terminals are connected to the points $\z_1$ and $\z_2$ of the robot.
The external forces, $\F_1$ and $\F_2$, respectively acting on $\z_1$ and $\z_2$, balance the force in the spring, and are given by the gradient of this potential:
$\F_1^T = \frac{\partial V(\z)}{\partial \z_1}$ and $\F_2^T = \frac{\partial V(\z)}{\partial \z_2}$.
The virtual force of the spring reads
\begin{equation}
\label{eq:spring}
\F^T = \frac{\partial V(\z)}{\partial \z}
\end{equation}
and the resulting joint actuation is thus given by
\begin{equation}
\label{eq:virtual_force_in_practice}
    \u = -(J_{\z_1}(\q)- J_{\z_2}(\q))^T \F.
\end{equation}
Any such spring is passive: taking the derivative of $V$,
\begin{subequations}
\begin{align}
    \dot{V} &= \dot{\bm{z}}_1^T \frac{\partial V}{\partial \bm{z}_1} + \dot{\bm{z}}_2^T \frac{\partial V}{\partial \bm{z}_2} \\
    &= \dot{\bm{z}}_1^T \bm{F}_1 + \dot{\bm{z}}_2^T \bm{F}_2,
\end{align}
\end{subequations}
we show that the spring satisfies \eqref{eq:PassivityDt}. Note the equality sign: the spring stores energy but does not dissipate energy.

\begin{figure}[t]
    \centering
    \includegraphics[]{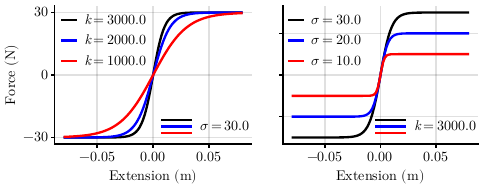}
    \caption{The nonlinear spring force $\F = \sigma \tanh(k|\z|/\sigma) \frac{\z}{|\z|}$
    for different values of the parameters $k$ and $\sigma$.}
    \label{fig:saturating spring}
\end{figure}

A linear spring takes $V(\z) = \frac{1}{2} k \z^T \z$, where $k$ is a positive scalar.
The corresponding virtual force is $\F = k \z$. Impedance controllers in robotics are often limited to linear springs. However, in applications, it is useful to adopt nonlinear springs, generated by nonquadratic potentials. For example, 
the potential energy $V(\z) = \frac{\sigma^2}{k} \ln (\cosh (k |\z|/\sigma))$ leads
to a virtual force of the form $\F = \sigma \tanh(k|\z|/\sigma) \frac{\z}{|\z|}$, as plotted in Fig. \ref{fig:saturating spring}. 
This `tanh-spring' models a useful saturation. In the context of robotic control, 
this component cannot apply a force greater than $\sigma$, which can be used to build mechanisms with a limited maximum force, for safer human-robot interaction. 
Similarly, the potential energy $k\frac{(\z^T \z)^p}{2p}$, $p>1$, makes the force
grow rapidly for $|\z| > 1$, which can be used to enforce constraints on robot motion.

\subsubsection*{Dampers}
Dampers resist motion by dissipating energy.
At any instant,
\begin{align}
    \label{eq:DampingCondition}
	\F^T \dot \z \geq 0
\end{align}
where $\F$ is the force exerted by the damper at its terminals,
as a function of their relative velocity $\dot \z$ 
and their relative displacement $\z$. 

A linear damper satisfies $\F= c\dot \z$, where $c \geq 0$ is its damping coefficient. 
Its braking action removes energy from the system, reducing oscillations. 
Nonlinear
dampers can be very useful for control purposes.
Saturating dampers $\F = c\tanh(\dot{\z})$ or localized dampers that are active only in a defined region $\F = c \mu(\z) \dot \z$, where $0 \le \mu(\z) \le 1$, can be used to implement many useful behaviors while remaining within the limits of a passive control action. 

Note that the action of a virtual damper on the robot is realized by \eqref{eq:virtual_force_in_practice} whenever $\z$ is associated to two generic points on the robot, $\z = \z_1 - \z_2$.

\subsubsection*{Inerters}
% Springs and dampers are most common because for a typical robot we have good measurements of positions and velocities impedance, so a spring or a damper can be implemented in feedback on a force-controller robot.
The inerter \cite{Smith2002} can be considered a generalized inertia/mass. It is used to add inertance to a mechanism. The force of a linear inerter is 
\begin{equation}
\F = m \ddot{\z}.
\end{equation}
where $m$ is the inertance. 
When one of the terminals is connected to ground, the inerter acts as a point-mass unaffected by gravity.

As virtual springs and dampers, the action of a virtual inerter is realized by \eqref{eq:virtual_force_in_practice} whenever $\z$ is associated to two generic points $\z = \z_1 - \z_2$.
In what follows, we avoid attaching virtual inerters to the robot directly, as it
is difficult to accurately determine the robot accelerations. %This is because the acceleration signal has more high frequency content than the position or velocity. 
We connect inerters to points that are internal to the virtual mechanism that characterize the controller, whose accelerations can be computed analytically. 
An equivalent choice is often made in many impedance controllers,
which shape only the stiffness and damping by
leaving the robot's inertia matrix unchanged (the `target' matrix corresponds to the nominal matrix) \cite{Averta2020, Lachner2022, Hogan2022, Lawrence1988}.

\begin{remark}
Like a spring, a nonlinear inerter could be defined 
using potentials $V: \mathbb{R}^3 \to \mathbb{R}$
through the relationship
\begin{equation}
\dot{\z}^T = \frac{\partial V(x)}{\partial x} \qquad \dot x = \F.
\end{equation}
where $x$ is an additional `state' variable, a generalized momentum.
This nonlinear inerter is passive if $V$ is lower bounded.
In this form, the inerter is an admittance
(force as input and motion as output).
The analogy with the spring follows by rewriting \eqref{eq:spring} as
\begin{equation}
\F^T = \frac{\partial V(x)}{\partial x} \qquad \dot x = \dot{\z}.
\end{equation}
We do not explore nonlinear inerters in this paper
to remain within a family of elements that can be easily represented
as impedances (motion as input and forces as output).
\end{remark}

\subsection{Virtual structures and implementation} \label{sec:VirtualStructures}

\begin{figure}[t]
    \centering
	\includegraphics[page=4, width=\columnwidth]{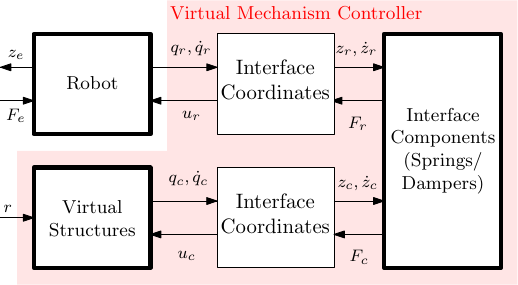}
	\caption{Virtual model controller block diagram. $(\F_e, \z_e)$ is an external port. $r$ represent external inputs to the controller. This is a general structure for any torque controller robot providing joint position and velocity feedback. 
    In the case of the virtual instrument controller of Fig. \ref{fig:VirtualInstrument}, described in detail in Section \ref{sec:SurgeryVM}, the `virtual structure' comprises of the rotating joints, and the mass/inertia of the virtual instrument itself, while the springs and dampers between the virtual instrument and robot are `interface components'.}
	\label{fig:BlockDiagram}
\end{figure}

Virtual springs, dampers, and inerters are combined into
virtual structures defined by a set of rigid bodies connected by rigid and prismatic joints. Extensions to higher-d.o.f. joints are straightforward.
Together, these compose the virtual mechanism that represents the
controller. The controller has dynamics, since
rigid bodies and joints introduce new states. The action of the controller
is ultimately determined by the interconnection with the robot, which is always
realized through springs and dampers. Fig. \ref{fig:BlockDiagram} provides an illustration of the virtual model controller and its interconnection with the robot. 

The virtual mechanism is a passive dynamic extension of the robot dynamics
(control by interconnection \cite{Ortega2008}). A virtual mechanism can model mechanical constraints, such as virtual fixtures \cite{Bowyer2014}, but its dynamics allow for the modeling of
more general dynamical constraints, virtual tools (as shown in Fig. \ref{fig:VirtualInstrument}), and even virtual copies of the robot (digital twin).
The geometry of the virtual structure combined with the tuning of the parameters of springs, dampers, and inerters shape the impedance \cite{Hogan1984} of the interaction port $(\F_e,\z_e)$, which captures the response of the robot to the external world.

We endow any virtual structure with mass/inertia/inertance such that its inertia matrix is nonsingular and it can be simulated as a second-order-dynamics.  
The implementation of the controller is captured in Fig. \ref{fig:BlockDiagram}.
Arrows indicate the flow of information. The robot and the virtual structure act as \emph{admittances}, taking force inputs and returning motion; the interface elements act as \emph{impedances} taking input from motion and returning the forces\footnote{Note that this uses the force-effort analogy, while there are compelling reasons to use a force-flow analogy, for a description see \cite[Section 3]{Smith2020}}.
To implement the virtual structures, they are simulated in real time, as 
\begin{equation}
    M_c(\q_c) \ddot{\q}_c + C_c(\q_c, \dot{\q}_c) \dot{\q}_c = \u_c
\end{equation}
where $\q_c$ represents the generalized coordinates of the virtual model controller,
$M_c$ is the combined inertia/inertance matrix (the latter due to inerters directly attached to the virtual structure), and $C_c(\q_c, \dot{\q}_c)$ models Coriolis/gyroscopic effects.

The interface elements are formed of spring and dampers, which ultimately depend only upon the position/velocity states of the robot (via the interface coordinates), and do not depend on accelerations.

\begin{remark}
If the many Jacobians need to be constructed to apply forces at a large number of points, these computations can form a significant fraction of the computational cost.
To improve performance in future implementations, explicit computation of Jacobians can be avoided entirely for serial robots or tree-structured robots by recursive algorithms, which propagate kinematics/velocities forwards through the structure from the base, and forces backward from the tip(s), while also computing the required joint actuation $\u$.
For an example, refer to the the backwards pass of the Recursive Newton Euler Algorithm in \cite{Featherstone2008}.
\end{remark}

\begin{remark}
Although we limit our discussion to fully actuated robots, virtual model control is still possible for underactuated robots. The mechanism must in that case be designed such that the un-actuated joints are not required to emulate the virtual forces on the robot. 
Indeed, in Fig. \ref{fig:OperationSpaceControl}, the force $\F_1$ does not require any actuation of the third joint.
Likewise, in a quadruped, the underactuation is due to the lack of rigid connection to the ground. However, one could design a virtual mechanism acting between any part of the body/legs of the robot \cite{Pratt1995b}.
\end{remark}

\section{The design of virtual mechanisms}
\label{sec:VirtualMechInControl}

\subsection{Virtual model control for reaching}
\label{sec:Reaching}

The advantages of designing a controller as a virtual mechanisms can be illustrated by revisiting the classical control of a `reach' motion, the basic primitive of any manipulator. Once gravity is compensated, the simplest virtual mechanism is given by a virtual spring and a virtual damper 
connected between the end effector and the target.
This corresponds to the classical PD control in the task space \cite{Lawrence1988}. 
It is straightforward to see that the adoption of nonlinear elements in this context enables important features. Consider a spring and a damper with 
nonlinear force-displacement characteristics
\begin{subequations}
\label{eq:nonlin_spring_and_dampers_1dof}
\begin{align}
\F_{\mathrm{spring}} & = \sigma_1 \tanh \left( \frac{k |\z|}{\sigma_1} \right) \frac{\z}{|\z|} \\
\F_{\mathrm{damper}} & = \sigma_2 \tanh \left( \frac{c |\dot{\z}| }{\sigma_2} \right) \frac{\dot{\z}}{|\dot{\z}|},
\end{align}
\end{subequations}
respectively. $\sigma_1>0$ and $\sigma_2>0$ capture the maximal magnitude of the saturation. In the linear regime of the hyperbolic tangent, $k>0$ and $c>0$ are 
stiffness and damping coefficients. 
The control action is still passive. The spring has energy $\sigma_1^2k^{-1}  \ln(\cosh(k|\z|/\sigma_1))$,
which is lower bounded. The damper dissipates energy with rate 
$\sigma_2 \tanh(c |\dot{\z}|/ \sigma_2) |\dot{\z}| $, which is always positive.
In contrast to the linear case, the saturation values $\sigma_1$ and $\sigma_2$ 
limit the static force that the end effector exerts on a generic static obstacle.
They also limit actuation efforts. 

\begin{figure}[t]
    \centering
    \includegraphics[width=\columnwidth]{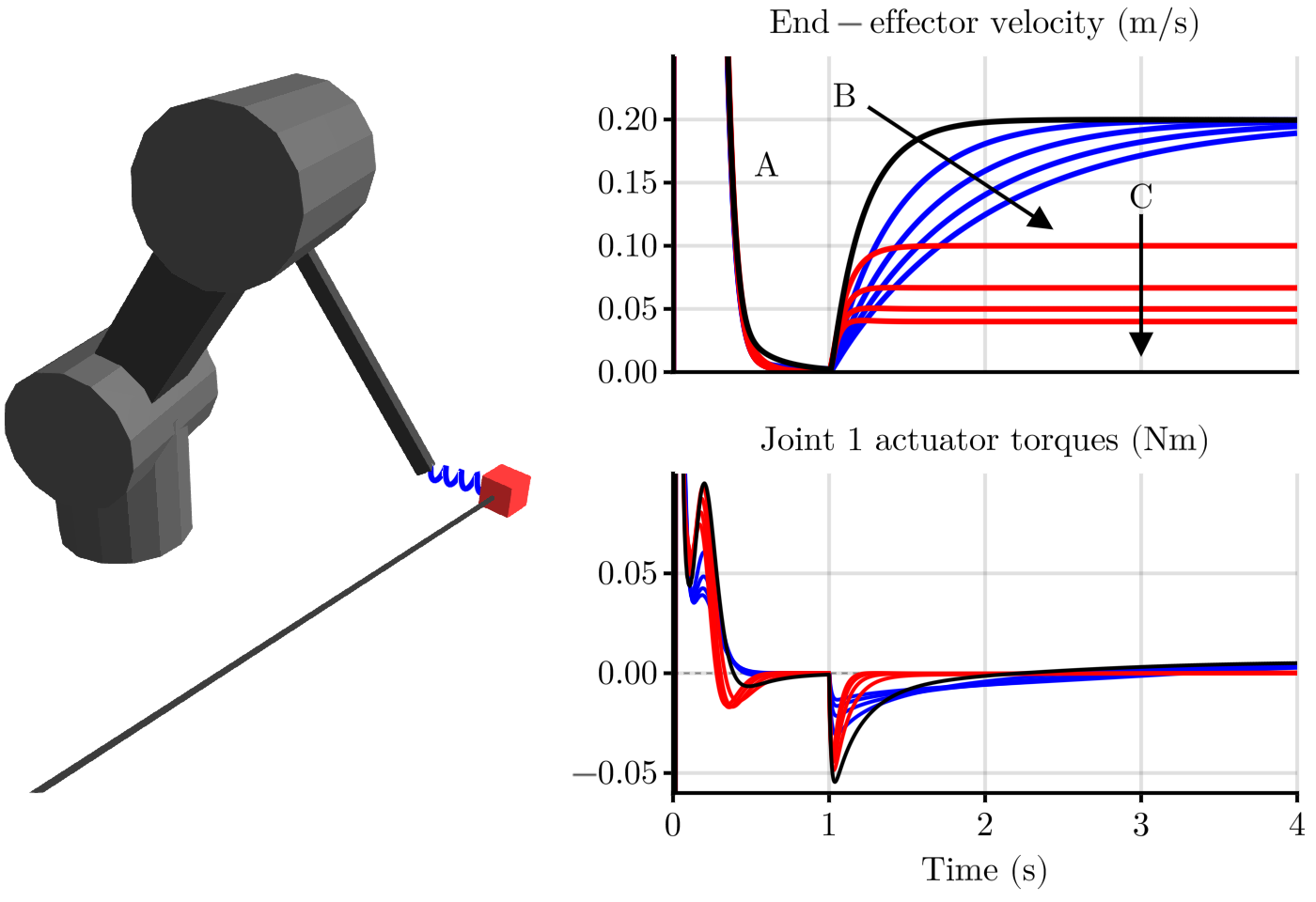}
    \caption{
     [Left] Virtual mechanism for reaching. Robot is constrained by a spring damper shown in blue to a virtual cart on a linear rail shown in red. 
     [Right] Simulated end-effector velocity in response to a step input in force (acting on the cart), with varying parameters.
     A) Initial transients, mostly identical. Nominal response shown in black.
     B) Varying response for increasing cart mass, blue. Steady state speed does not vary, but acceleration slows.
     C) Varying response for increasing cart damping, red. Top speed decreases but similar acceleration.
     }
    \label{fig:mass_cart}
\end{figure}

The more advanced exemplary virtual mechanism of Fig. \ref{fig:mass_cart} allows for refined control over smoothness and peak velocity of the reaching motion. Linear spring and dampers are now attached between the end effector and virtual moving cart. The cart is pulled by a constant force, experiences some drag due to friction, and has mass, which leads to a dynamic controller. 

This design has several useful features.
It separates the tasks of constraining the robot and driving motion along the line: the spring-damper components between the damper and the cart regulate the constraint, while forces/components applied to the cart regulate the reaching motion. 
By driving the cart with a constant force pointed in the direction of the final desired position, rather than as by a reference \emph{position} that moves as a function of time, we achieve a safer form of path following.  
Consider an obstruction such as a person in the path of the robot: the end-effector and person collide and the cart continues to move forwards only until equilibrium is reached between the driving force and the constraint-spring. Thus, even for a stiff spring ensuring a strong constraint to the path, we can achieve compliant behaviour along the direction of motion.
In addition the velocity of the robot end-effector can be regulated by tuning the friction applied to the cart: neglecting the robot's friction,  the steady state velocity can be determined by balancing the cart's friction with the applied driving force. 
Finally, robot fast accelerations can be moderated by increasing the inertance of the cart.
The cart also has the effect of filtering the driving force: a step change in driving force due to a sudden change of reference (or noise) results in smooth driving torques.
For a numerical example of these effects, see Fig. \ref{fig:mass_cart}.

The above considerations illustrate how the design of virtual mechanisms carries significant insights on the control action.
While equivalent behaviour can be achieved with impedance/force control, virtual mechanisms provide a framework for reasoning about controller design decisions, particularly in cases more complicated than this pedagogical example.
By relying on physical intuition, we can make predictions about the robot reactions to obstacles, the velocity peak of the end effector, and the filtering effects of the controller dynamics.
It is also easier to understand how the parameters of the virtual mechanism will
affect the control performance. For example, it is straightforward to imagine that a larger cart mass will induce a stronger lowpass effect on the pulling
force, as shown in Fig. \ref{fig:mass_cart}.  Similarly, a larger spring stiffness may lead to higher peak forces.

% One other research area in which nonlinear, operation-space energy shaping has been explored is the field of virtual fixtures, sometime known as active constraints.
% Often used in robotic surgery, these are energy shaping force fields that guide the surgeon towards or away from certain features/obstacles, for example constraining the end-effector to a certain trajectory, region or planned motion \cite{Cai2022, Chen2016, Nasseri2014, Selvaggio2018, Bowyer2014}. 
% \todo{What else to say here?}

\subsection{Virtual model control for laparoscopic surgery} 
\label{sec:SurgeryVM}

\begin{figure}[t]
    \centering
    \includegraphics[page=2, width=\columnwidth]{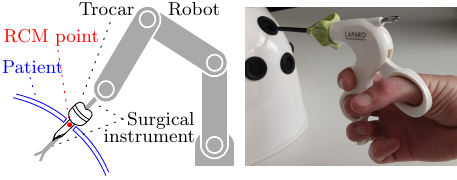}
    \caption{
    [Left] Graphical illustration of the main robotic assisted laparoscopic surgery tasks. [Right] Manual laparoscopic instrument and surgical phantom, for comparison.}
    \label{fig:RMIS}
\end{figure}

Virtual model control appears particularly effective on complex tasks.
Here we consider robotic surgery, specifically
laparoscopic surgery, as shown in Fig. \ref{fig:RMIS}. The controller has two main objectives:
(i) the instrument must pass through a small aperture in the patient's body, often achieved by maintaining the tool over the so-called Remote Center of Motion (RCM), a fixed point in space;
(ii) the `tip' of the instrument must track the reference indicated by the surgeon.

Fig. \ref{fig:VirtualInstrument} provides a graphical representation of a virtual mechanism for laparoscopy. After gravity compensation, a `virtual instrument' is virtually connected to the actual laparoscopic instrument attached to the robot, constraining and driving its motion. 
It is composed of the virtual instrument itself---a rigid body structure with a spherical joint located at the RCM and a co-located prismatic joint---and a set interface spring-damper pairs.
By construction, axis of the virtual instrument always lies on the RCM. The interface springs pull the instrument towards the virtual instrument.
At equilibrium, the instrument and the virtual instrument will be co-located, satisfying the RCM constraint. At the same time, the spring and damper between the end effector at the tip of the instrument and the reference position $\bm{r}$ will cause the end effector to track the surgeon's motion.

\begin{figure}[t]
    \centering
    \setlength{\columnsep}{0pt}
    \includegraphics[width=110px, height=110px]{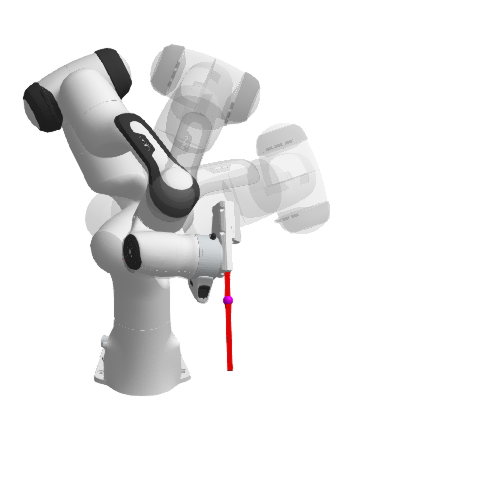}
    \includegraphics[width=110px, height=110px]{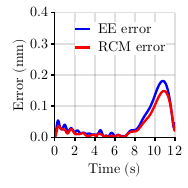}
    \vspace{-.5cm}
    \caption{
         Null space motion from a starting pose (solid) and middle/final poses (translucent) for a constant end-effector/RCM position with a 1Nm torque applied at J1 for 12 seconds. End-effector and RCM error remain low ($<\SI{1}{mm}$) while moving in the null-space. 
     }
    \label{fig:cool_virtual_instruments}
\end{figure}

\begin{figure}[t]
    \centering
    \includegraphics[width=220px, height=110px]{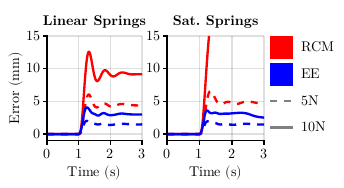}
    \vspace{-.5cm}
    \caption{
         Tracking error when a disturbance force is applied at the RCM. The disturbance force is either 5N or 10N.
         The two graphs show results with linear springs between the virtual-instrument, or saturating springs that saturate at 10N. 
         For small disturbance forces, behaviour is similar, but for large disturbances we see compliant behaviour for a large disturbance.
         In all cases the end-effector error remains low.
     }
    \label{fig:cool_virtual_instruments2}
    \vspace{-.2cm}
\end{figure}

Compared to PD based approaches, our approach does not require inverse kinematics, and therefore gracefully handles redundancy in the robot structure, producing an unconstrained `null space' (see Fig. \ref{fig:cool_virtual_instruments}).
The controller is essentially identical for different robot morphologies.
The spring placement allows us to stiffen the robot's behaviour at the RCM independently of the end-effector: compared to energy shaping, the mechanical analogy of the virtual mechanism makes design alterations intuitive, guided by our understanding of the physics of the system.
The plots of Fig. \ref{fig:cool_virtual_instruments2} shows the effect of replacing the linear springs between the robot and virtual instrument with saturating springs (described in Fig. \ref{fig:saturating spring}), when faced with a step disturbance force at the RCM.
The saturating spring allows a compliant behaviour at the RCM for large disturbances.

So long as the saturation force of the springs is larger than the standard required operational forces, typical behaviour is minimally affected.
These springs could allow surgical staff to manually override the controller by pulling with a large enough force, and we use a similar design for robot assisted drilling in \cite{Larby2025}.
This design change is enabled by the virtual mechanism analogy.

To implement virtual model controllers no additional sensing beyond torque actuated joints with position/velocity feedback is required, and \textit{dynamic} tuning/filtering can be achieved by changing inertance/damping of virtual instrument.
Importantly, because the virtual mechanism is passive, the system is stable for many types of model uncertainty (parametric or dynamic perturbations preserves the passivity of the robot, and external interactions as long as the environment remains passive).
If the actuators are capable of faithfully producing the desired control forces, then the system is stable regardless of the parameterization of the controller (although in practise there are upper limits on stiffness and damping that can be realized).
This helps bridge the sim-to-real gap, as all models are uncertain to a degree, and additionally the system has many unmodified external interactions.
\begin{remark}
The virtual mechanism presented here is based on the 
`Virtual Instrument' presented in  \cite{Larby2022b}, but has a simpler structure.
This is due to the tuning method discussed in the next section, which
allows for more general virtual mechanisms than \cite{Larby2022b}.
\end{remark}


\section[Optimal virtual mechanisms: formulation and numerical solution]{Optimal virtual mechanisms: \\ formulation and numerical solution} 
\label{sec:OptimalVirtualMechanisms}

\subsection{Input/output performance}
\label{sec:CostFunctions}

Once the virtual mechanism is designed to fulfill the task objectives,
its parameters can be adjusted for performance optimization.
Despite the intuitive link between design (the mechanism components) and performance (the task), and despite the resilience of the controller to a wide range of parameters and model uncertainties, it is still desirable to fine-tune the performance of the system by considering the response to
external disturbances and noise.

\begin{figure}[t]
    \centering
    \includegraphics[width=0.6\columnwidth]{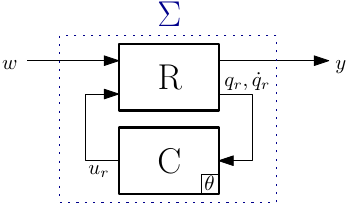}
    \caption{System $\Sigma$ from $\bm{w} \rightarrow \bm{y}$.}
    \label{fig:genericHinf}
\end{figure}

In what follow we consider the robot as an open system, as shown in Fig. \ref{fig:genericHinf}.
The robot is represented by the upper block, $R$, and the virtual model controller is represented by the lower block, $C$. Performance inputs $\w$ model exogenous perturbations to the robot that arise from the interaction with the working environment, such as contact forces. Performance outputs $\y$ capture the behavior of interest, that is, displacements and velocities (dynamically weighted) that are relevant to the task.
The virtual parameters of the controller are represented by the vector ${\bm{\theta}}$, whose elements are greater than or equal to zero, modeling stiffness and damping coefficients, and inertia and geometric parameters such as length and weight of virtual links. 

Fig. \ref{fig:genericHinf} represents a classical setting of linear robust control theory \cite{Zhou1996}, whose objective is to design the controller to guarantee specific closed-loop performance from $\w$ to $\y$.  In this paper, the controller is restricted to a family of (nonlinear) virtual mechanisms, parameterized
by ${\bm{\theta}}$, and the goal is to find the parameters ${\bm{\theta}}$ that optimize a combined $\mathcal{L}_2$ and $\mathcal{L}_\infty$ performance from $\w$ to $\y$. Namely, considering the splitting of the signals $\w = (\w_1,\w_2)$ and $\y = (\y_1,\y_2)$, 
we want to optimize the aggregated cost 
\begin{align}
    \min_{\bm{\theta}} \max_{\w} & \ L({\bm{\theta}}, \w) \nonumber\\
    \mbox{ where} &\ L({\bm{\theta}}, \w) = \frac{\parallel\!\y_1\!\parallel_2}{\parallel\!\w_1\!\parallel_2} + \frac{\parallel\!\y_2\!\parallel_\infty}{\parallel\!\w_2\!\parallel_\infty}
    \label{eq:performance metric}
\end{align}
with the standard meaning $\parallel\!\w_1\!\!\parallel_2 = \left(\int_0^\infty \! \w_1(t)^T \w_1(t) dt\right)^{\frac{1}{2}}$ and
$\parallel\!\w_2\!\!\parallel_\infty = \sup_{t \geq 0} |\w_2(t)|$, assuming boundedness of the input signals in their respective domains.

By maximizing over $\w$, we are looking for the `worst' case, that is, the largest amplification factor from $\w$ to $\y$ (the so-called gain of the system). This can be used to minimize the effect of exogenous disturbances or tracking errors.
The ratio between $\parallel\!\y_1\!\parallel_2$ and $\parallel\!\w_1\!\parallel_2$ in \eqref{eq:performance metric} is based on the \emph{time averages} of the performance signals $\w_1$ and $\y_1$.  Therefore, it looks at the long-term effect that a perturbation may have on the robot
(this corresponds to the $\mathcal{H}_\infty$ system norm). 
In contrast, the ratio between $\parallel\!\y_2\!\parallel_\infty$ and $\parallel\!\w_2\!\parallel_\infty$ looks at \emph{peaks}, and can be used to analyze short-range phenomena, such as the peak displacement caused by a contact force of given magnitude (this corresponds to the $\mathcal{L}_1$ system norm). 
We emphasize that the choice of the relevant input/output pairs $(\w_1,\y_1)$
and $(\w_2,\y_2)$ is going to be determined by the use case.

By optimizing over ${\bm{\theta}}$ in \eqref{eq:performance metric}, we constrain the control action within a predefined set of virtual mechanisms while shaping the response of the controlled robot from $\w$ to $\y$, i.e. its closed-loop impedance $\w$ to $\y$. In this sense, \eqref{eq:performance metric} extends the classical framework of interconnection and damping assignment \cite{Ortega2008} to impedance control \cite{Hogan1984} and performance optimization. 

\subsection{Tractable optimization}

The min-max form of \eqref{eq:performance metric} is a notoriously difficult type of optimization problem \cite{Hsieh2021}. In contrast to linear control theory \cite{Zhou1996}, \eqref{eq:performance metric} is far from tractable in the nonlinear and structured setting of virtual mechanisms, as it requires infinitely many probing signals $\w$. To solve this optimization, we need to build a sequence of approximations that leads to a well-defined, tractable optimization problem. 
We propose two approaches.

\underline{\emph{Sampling}}: select a \emph{finite} family, $\mathcal{W}$, of perturbing signals $\w$ that are relevant to the engineer. This family could include perturbations that are significant for the specific task or that are known to cause problems. Each sampled signal identifies a \emph{scenario} that the user has identified as relevant. In this case \eqref{eq:performance metric}  reduces to 
\begin{equation}
\label{eq:performance metric_sampled}
\min_{{\bm{\theta}}} \max_{\w\in \mathcal{W}}  \ L({\bm{\theta}}, \w) 
\end{equation}

\underline{\emph{Adversarial}}: alternatively, an adversarial approach could be used to find relevant probing signals that induce poor performance. 
By maximizing the cost over \emph{parametrized} perturbing signals $\w(\omega)$,
where $\omega$ is the parameter vector,
we can iteratively grow the family of probing signals (scenarios) $\mathcal{W}_k$, $k \in \mathbb{N}$, by adding those that represent poor operating points of the system. In such a case, the optimization \eqref{eq:performance metric} becomes a two-step optimization of the form: 
\begin{subequations}
\begin{equation}
\label{eq:performance_metric_adversarial}
\min_{{\bm{\theta}}_k} \max_{\w\in \mathcal{W}_k}  \ L({\bm{\theta}}_k, \w) 
\end{equation}
\begin{equation}
\label{eq:performance_set_adversarial}
    \mathcal{W}_{k\!+\!1} = \mathcal{W}_k \cup \{ \w(\overline{\omega})\} \mbox{ for }
    \overline{\omega} = \mathrm{argmax}_{\omega} \, L({\bm{\theta}}_k, \w(\omega)) \vspace{2mm} 
\end{equation}
\end{subequations}
starting from a given initial $\mathcal{W}_0$ and iterating over $k$.
Termination could be enforced after a pre-defined number of steps or 
by considering a bound on the residual improvement.

At the cost of suboptimality, these two approaches simplify the optimization since the inner maximization is now reduced to a finite set of signals (possibly growing). 
The outer minimization can be solved via gradient descent, as discussed in the next section. Even if suboptimal,
Figs. \ref{fig:L2GainSampled}-\ref{fig:LinfGainAdversarial} show that
a sufficiently dense set $\mathcal{W}$ for the sampling approach, or a few iteration steps for the adversarial one, provides neat performance results.

\subsection{Solution via algorithmic differentiation} \label{sec:Tuning}

The minimization problem above can be solved by algorithmic differentiation, \cite{Rackauckas2021a,Ma2021}. The optimal vector ${\bm{\theta}}$ can be computed by gradient descent through ODE simulations. 
%However, this can be computationally expensive. Naively using automatic differentiation on a cost function calling an ODE solver internally can result in errors, so 
We make use of the Julia package \emph{SciMLSensitivity.jl} \cite{Rackauckas2021}, which integrates with a modular algorithmic differentiation ecosystem in Julia to provide differentiation rules for many ODE solvers in forward and backward modes.
In this paper, we use automatic differentiation in the forward mode, which becomes slower to compute the gradient as the number
of parameters in ${\bm{\theta}}$  scales, but is fast for a small number
of parameters and can be used with minimal code adaptation. 

A detailed description of automatic differentiation algorithms is beyond the scope of this paper. For parameterized ODE of the form $\dot{\x} = f(\x,{\bm{\theta}})$, we recall that forward 
automatic differentiation computes the
gradient with respect to ${\bm{\theta}}$ by solving forward in time the 
extended system
\begin{equation}
\label{eq:fadn}
\frac{d}{dt} \begin{bmatrix}
\x \\
\frac{\partial \x}{\partial {\bm{\theta}}} \\
\end{bmatrix}
=
\begin{bmatrix}
f (\x, {\bm{\theta}})  \\
\frac{\partial f(\x, {\bm{\theta}})}{\partial \x} \frac{\partial \x}{\partial {\bm{\theta}}} + \frac{\partial f(\x, {\bm{\theta}})}{\partial {\bm{\theta}}}
\end{bmatrix}.
\end{equation}
The gradient of $\x$ at a generic time $t$ is thus obtained by initializing
\eqref{eq:fadn} with a suitable selection of
$\frac{\partial \x}{\partial {\bm{\theta}}}$ at time $0$. 
Thus, to use automatic differentiation, we need to reformulate 
our optimization problem as a (differentiable) ODE of the form $\dot{\x} = f(\x,{\bm{\theta}})$, possibly paired to some readout function $h(\x(t))$, where $h$ (differentiable) nonlinear mapping of the vector $\x$ at time $t$. Together, the ODE and the readout function must capture the complete dynamics of the robot and the virtual mechanism. They must also contain additional state variables to compute the aggregate cost $L$ in \eqref{eq:performance metric}. 

Transforming robot and virtual mechanism's dynamics into a first-order ODE is standard. For the aggregate cost $L$, we proceed as follows.
\begin{itemize}
\item For the $\mathcal{L}_2$ part, 
we replace the indefinite integral
$\int_0^{\infty} \dots dt$,
with the definite one $\int_0^{T} \dots dt$,
where $[0,T] \subset\mathbb{R}$ is the chosen simulation interval.
Then, we compute this integral by introducing two new state variables,
${c}_{\w_1}$ and ${c}_{\y_1}$, and the ODE formulation
\begin{align}
    \dot{c}_{\w_1} &= \w_1^T \w_1 \nonumber \\
    \dot{c}_{\y_1} &= \y_1^T \y_1 \nonumber 
\end{align}
The value of the $\mathcal{L}_2$ part of the cost 
is thus given by 
$\sqrt{\frac{c_{\y_1}(T)}{c_{\w_1}(T)}}$. In practice, we use
$\sqrt{\frac{c_{\y_1}(T)}{\epsilon+c_{\w_1}(T)}}$ to avoid issues of division by zero due to possible non-zero output transients $\y_1$ for small/zero inputs $\w_1$, by introducing $\epsilon=10^{-6}$ in the denominator.
\item For the $\mathcal{L}_\infty$ part, we introduce two new states,
${c}_{\w_2}$ and ${c}_{\y_2}$ and we consider the relaxation based on the ODE
\begin{align}
	\tau \dot{c}_{\w_2} &= \max({c}_{\w_2}, |\w_2|) - {c}_{\w_2} \nonumber \\
	\tau \dot{c}_{\y_2} &= \max({c}_{\y_2}, |\y_2|) - {c}_{\y_2} \nonumber 
\end{align}
For sufficiently small time constant $\tau >0$, ${c}_{\w_2}(T)$ and ${c}_{\y_2}(T)$  tend towards the infinity norm of signals $\w_2$ and $\y_2$, within the simulation interval $[0,T]$.
The value of the $\mathcal{L}_\infty$ part of the cost  is thus given by $\frac{{c}_{\y_2}(T)}{{c}_{\w_2}(T)}$. Again, in practice, 
we use $\frac{{c}_{\y_2}(T)}{\epsilon + {c}_{\w_2}(T)}$.
% \item Finally, each use of the $\max$ function, also the one in \eqref{eq:performance metric}, is replaced by a softmax function to get a continuous gradient.
\end{itemize}

\section[Optimal virtual mechanisms: performance signals and parameter tuning]{Optimal virtual mechanisms: \\ 
performance signals and parameter tuning} 

\subsection{Optimal PD control of a cart}
\label{sec:PDcart}

\begin{figure}[t]
	\centering
	\includegraphics{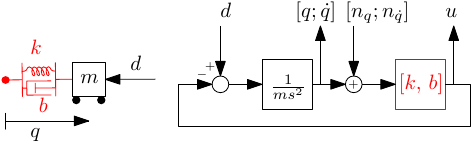}
	\caption{Illustration of the 1DOF system, with 2 controller parameters and block diagram with exogenous inputs/outputs.}
	\label{fig:1dof}
\end{figure}

We illustrate the controller optimization on the simplest example of 
a PD control of a cart. From Fig. \ref{fig:1dof},
the equations of cart and controller read
\begin{subequations}
\label{eq:1dofDynamics}
\begin{align}
	m \ddot q & = u + d \\
        u & =  -k (q + n_q) +  b (\dot{q} + n_{\dot q}),
\end{align}
\end{subequations}
where $q$ and $\dot{q}$ are position and velocity of the cart, 
$d$ is an external force acting on the cart, and 
$\n_q$ and $\n_{\dot{q}}$
are disturbances affecting position and velocity measures. 
$m$ is the mass of the cart and $k$ and $b$
are the proportional and derivative parameters of the controller, 
which correspond to virtual mechanical stiffness and damping, respectively.
These are collected in the parameter vector ${\bm{\theta}}$.

For the optimal tuning, we select the 
performance input as weighted disturbance
and the performance output as a weighted readout of cart position and control force. 
Namely, 
\begin{equation}
        \begin{bmatrix}
		d  \\
            n_q \\
		n_{\dot q}
	\end{bmatrix}   = W_w \w \ \mbox{ and } \ 
        \y   = W_y         
        \begin{bmatrix}
		q  \\
            \dot{q} \\
		u
	\end{bmatrix} 
\label{eq:performance_input_output}
\end{equation}
where
\begin{eqnarray}
	W_w = \begin{bmatrix}
		0.01 & 0 & 0 \\
		0 & 0.1 & 0 \\
		0 & 0 & 5 \\
	\end{bmatrix}
	&
	W_y = \begin{bmatrix}
		100   & 0 & 0 \\
		0 & 0 & 0 \\
		0 & 0 & 1
	\end{bmatrix}
\end{eqnarray}
are specific (static) weight matrices.
%A block diagram representation is in Fig. \ref{fig:Weighting}.

With this selection of the performance pair $(\w,\y)$, we can optimize the control parameters to reduce the controlled cart
sensitivity to measurement noise and to external forces acting on it.
For instance, by tuning the control parameters to minimize the ratio between $\y$ and $\w$,
we make the controlled cart less sensitive to these disturbances. 
The controlled cart will attempt to hold its position even if perturbed by an external force, while mitigating the control effort. 
% \begin{figure}[htbp]
% 	\centering
% 	\includegraphics{figures/Weighting.pdf}
% 	\caption{Block diagram of the input/output weighting and signals}
% 	\label{fig:Weighting}
% \end{figure}

\begin{figure*}[htbp]
    \includegraphics{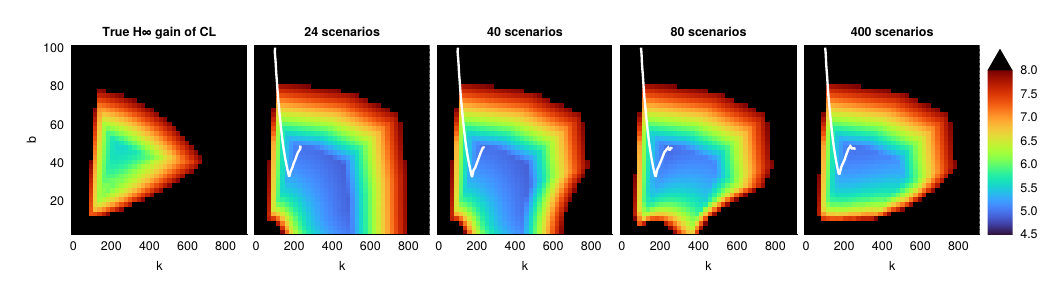}
    \vspace{-.8cm}
    \caption{Controlled cart $\mathcal{L}_2$ cost estimation with sampled scenarios. The white line shows the controller parameters while the optimization converges.}
    \label{fig:L2GainSampled}
    \vspace{-.4cm}
\end{figure*}

\begin{figure*}[htbp]
    \includegraphics{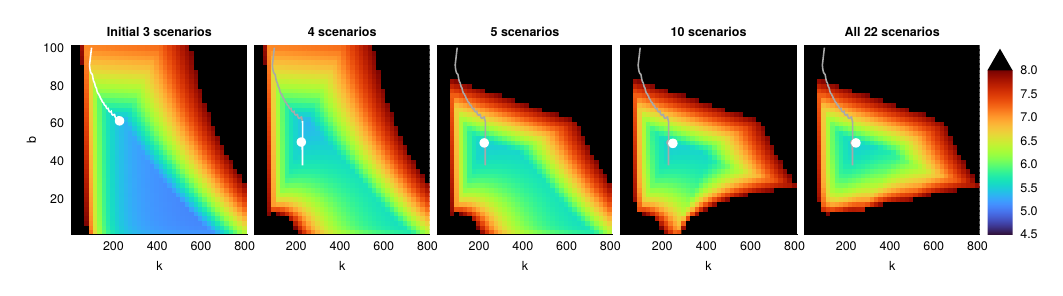}
    \vspace{-.8cm}
    \caption{Controlled cart $\mathcal{L}_2$ cost estimation with adversarial scenarios. The white/line shows the controller parameters while the optimization converges.}
    \label{fig:L2GainAdversarial}
    \vspace{-.4cm}
\end{figure*}

\begin{figure*}[htbp]
	\includegraphics{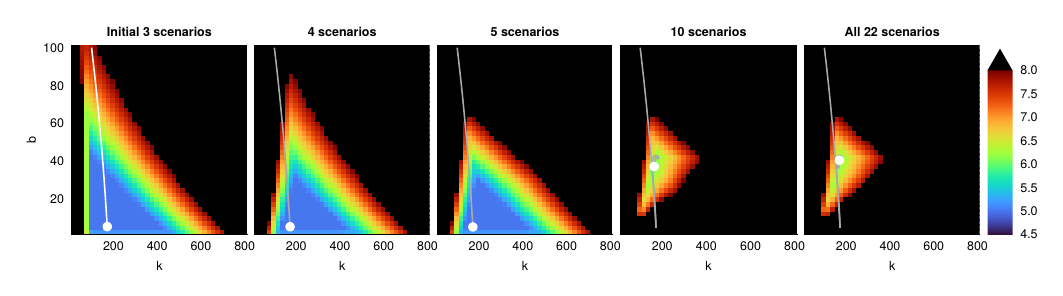}
	\vspace{-.8cm}
	\caption{Controlled cart $L_\infty$ cost estimation with adversarial scenarios.}
	\label{fig:LinfGainAdversarial}
 \vspace{-.4cm}
\end{figure*}

As \underline{first case} we focus on the $\mathcal{L}_2$ cost, by setting
\begin{equation}
\label{eq:L2cost}
\w_1 = \w \,,\ \y_1 = \y \,,\ \w_2 = 0 \,,\ \y_2 = 0 \,,
\end{equation}
in \eqref{eq:performance metric}.
We start with the \emph{sampling approach}. We create a set of scenarios $\mathcal{W}_N$ given by sampled sinusoidal signals of increasing frequency
\begin{equation}
	\label{eq:1dofSinusoid}
        \eta_i \sin(\Omega_i t) \quad \mbox{ for }
	\Omega_i = 0.1 \left(\frac{500}{0.1}\right)^\frac{i-1}{N-1}
 \!\!\! \mbox{and }  i \in \{1,\dots,N\},
\end{equation} 
with $\eta_i$ randomly sampled from the space of 3D unit vectors.
Frequencies are sampled exponentially between $0.1$rad/s and $500$rad/s.
The lower frequencies are sampled more densely, as this is where the dynamics of interest occur.

Fig. \ref{fig:L2GainSampled} shows a comparison between the true gain $\mathcal{L}_2$ obtained by $H_\infty$ synthesis \cite{Zhou1996}, and the result of our optimization,
as the size of the set of scenarios grows. 
As a greater number of frequencies are sampled, the cost landscape more closely resembles the `true' cost. The white lines represent the trajectory of the parameters $k$ and $b$ as the optimization 
converges\footnote{We have used the ADAM optimizer with an initial step size of $1.0$}.

Sinusoidal probing is effective here because the controlled cart is a linear system.
In general, for nonlinear robots controlled by nonlinear virtual mechanisms we would need a richer sampling space. In practice, we cannot sample
infinitely many input signals $\w$, but we can choose them
to cover a representative set of scenarios that the robot will face.
That is, we can design our probing space in relation to the task.
We also observe that,
while the cost of the computation scales linearly with the number of sampled
signals, the optimization is parallelizable. If required, optimizing over large scenarios can be made feasible by allocating additional computational resources.

The \emph{adversarial approach} provides an alternative to dense sampling, dramatically reducing the number of scenarios. 
Fig. \ref{fig:L2GainAdversarial} shows the optimization based on the adversarial approach. 
The two steps of optimization \eqref{eq:performance_metric_adversarial} and
\eqref{eq:performance_set_adversarial} run on scenarios
parameterized by four elements $\omega^i = [\,\omega_1^i , \omega_2^i , \omega_3^i ,\omega_4^i \,]^T$, specifically
\begin{equation}
\Omega_i = \omega_1^i \quad \eta^i = [\,\omega_2^i , \omega_3^i ,\omega_4^i \,]^T,
\end{equation}
where
$\omega_1^i$ is sampled in the range $[0, 500]$\footnote{\dots by taking the exponential of a uniformly sampled parameter in the range $[0, \log(500)]$, so that there is some bias towards lower frequencies.}, and $[\,\omega_2^i , \omega_3^i ,\omega_4^i \,]$ is sampled from the standard normal distribution.
An adversarial optimizer is initialized\footnote{An ADAM optimizer with initial step size $2.0$} and the `worst case' probing signal is obtained by approximating \eqref{eq:performance_set_adversarial} via 500 iterations (the controller does not change in this phase). 
Then, the controller adjustment \eqref{eq:performance_metric_adversarial} is resumed. And so on.
Fig. \ref{fig:L2GainAdversarial} illustrates the cost function for increasing
scenario sizes, starting from a very small setting of three signals. Each new scenario is optimally chosen by \eqref{eq:performance_set_adversarial}.

The summary in Table \ref{tab:L2Results} and the visual comparison between Fig. \ref{fig:L2GainAdversarial} and
Fig. \ref{fig:L2GainSampled} show that the adversarial case
is in closer agreement with the true cost landscape (top left of Fig. \ref{fig:L2GainSampled}). Remarkably, this is achieved with only 22 scenarios, a lot less than the 400 scenarios of the sampling case.
As the computational cost scales linearly with the number of scenarios, the adversarial approach offers an order-of-magnitude improvement in the tuning time.

\begin{table}[t]
	\centering
    \caption{Controlled cart parameters and estimated $\mathcal{L}_2$ cost.}
	\label{tab:L2Results}
	\begin{tabular}{|l|lll|}
            \hline
		& Cost   & k      & b     \\
            \hline
		True gain     & 5.5362 & 237.68 & 50.0  \\
		24 Scenarios  & 4.9681 & 231.11 & 48.51 \\
		40 Scenarios  & 4.9675 & 230.91 & 48.56 \\
		80 Scenarios  & 4.9299 & 259.91 & 47.61 \\
		400 Scenarios & 4.9692 & 254.18 & 48.26 \\
		Adversarial   & 5.4942 & 244.24 & 49.63 \\ \hline
	\end{tabular}
    \vspace{1pt}
\end{table}

% Between the 4 sampling scenarios of Fig. \ref{fig:L2GainSampled} the final controller and costs are similar. 
% This is perhaps because the direction of the input signals $\eta$ is not sampled more densely as the number of scenarios increases. 
% Perhaps surprisingly, they all converge to a similar region of $k, b$, close to the optimal values.

As \underline{second case} we now consider the $\mathcal{L}_\infty$ cost, by setting
\begin{equation}
\w_1 = 0 \,,\ \y_1 = 0 \,,\ \w_2 = \w \,,\ \y_2 = \y \,,
\end{equation}
in \eqref{eq:performance metric}. 
In this case, we use directly adversarial optimization. For simplicity, 
we use the parameterization \eqref{eq:1dofSinusoid} as in the $\mathcal{L}_2$ case, even if
sinusodial signals are not necessarily the best choice to calculate the $\mathcal{L}_\infty$ cost. The results are illustrated in
Fig. \ref{fig:LinfGainAdversarial}, which shows significant similarities and differences compared to the cost landscape of Fig. \ref{fig:L2GainAdversarial}.
The optimal proportional gain $k$ is smaller, while the derivative gain 
$b$ remains unchanged between the two cases.
In contrast to the $\mathcal{L}_2$ case, we cannot offer here 
a comparison with the `true' $\mathcal{L}_\infty$ to $\mathcal{L}_\infty$ gain, as 
this control synthesis problem is still unsolved, even in the linear setting.

\subsection{Optimal Virtual Model Control for Laparoscopic Surgery} \label{sec:SurgeryExample}

We demonstrate the applicability of the optimization approach to useful robotic systems, which have several degrees of freedom, and a more complex virtual-mechanism controller, by applying the tuning procedure to the virtual instrument controller for surgery described in \ref{sec:SurgeryVM}, with an additional virtual joint damper on J1 and on J4 (i.e. an additional torque proportional to  $\dot{q}_1$ applied by $u_1$, and on J4) to arrest null space motion in the simulation/tuning (which features no modelled joint-damping).

The robot manipulator kinematics is that of Franka Emika Research 3. We use the inertial parameters determined in \cite{Gaz2019}.
The virtual instrument is given an inertia of $0.05I \,\mathrm{kg\, m}^2$  and an inertance of $1.0 \,\mathrm{kg}$ by attaching a linear inerter between its centre and ground\footnote{This is equivalent to endowing a point mass of 1kg in zero gravity.}.
For simplicity, these parameters are set and will not be tuned in this paper.
There are $3$ virtual springs and $5$ virtual dampers to optimize, which leads to a vector of control parameters ${\bm{\theta}}$ of eight elements.

We focus on the $\mathcal{L}_2$ norm, adopting again \eqref{eq:L2cost}.
To enforce non-negativity of stiffness and damping parameters, 
 we introduce a nonlinear mapping $\phi$ between optimization parameters ${\bm{\theta}}$ and stiffness/damping values $\hat{\bm{\theta}}=\phi({\bm{\theta}})$.
Specifically, we take $\hat{\bm{\theta}}_i=\exp({\bm{\theta}}_i)$ if the parameter is a stiffness, and $\hat{\bm{\theta}}_i=\frac{1}{100}\exp{{\bm{\theta}}_i}$ if the parameter is a damping coefficient.
We choose the exponential function as it has no discontinuities and provide a suitable
parameter normalization for the optimizer. We also consider the addition of a regularization term 
\begin{equation}
    \label{eq:regularization}
    \sum_i \max(0, |{\bm{\theta}}_i| - 3000)^2
\end{equation}
to limit the maximum stiffness and damping values.
We penalize components of ${\bm{\theta}}$ above $\ln{}(3000)$, as robot actuators cannot realize excessive stiffness / damping. This corresponds to a soft limit on the stiffnesses at $3000$ and the damping coefficients at $30.0$\footnote{This is a crude way to avoid reaching the performance limits of our robot. A better identification of the bandwidth limits of the joints is needed to develop a more principled approach, beyond the scope of this paper.}. The corresponding values pre and post optimization are shown in Table \ref{tab:parameters}. The exact placement of the springs is shown in Fig. \ref{fig:VirtualInstrument}. The following abbreviations are used: End-Effector $\rightarrow$ EE, Virtual Instrument $\rightarrow$ VI, reference $\rightarrow$ ref, Joint 1 $\rightarrow$ J1.

\begin{table}[t]
    \centering
    \caption{Initial and tuned parameters (rounded).}
	\label{tab:parameters}
    \begin{tabular}{|l?{1.5pt}lrl?{1.5pt}lrl|}
        \hline
        & \multicolumn{3}{c?{1.5pt}}{Initial} & \multicolumn{3}{c|}{Final} \\ 
        Description & \,\,\,\,${\bm{\theta}}$ & $\hat{\bm{\theta}}$ && $\,\,\,\,{\bm{\theta}}$ & \qquad$\hat{\bm{\theta}}$ &  \\
        \hline
        EE-ref stiffness &  $6.91$ & $\num{1000}$ & \hspace{-3mm}$\unit{N/m}$ & $8.02$ & $\num{3056.4}$ & \hspace{-3mm}$\unit{N/m}$   \\
        Base-VI stiffness & $6.91$ & $\num{1000}$ & \hspace{-3mm}$\unit{N/m}$ & $7.22$ & $\num{1368.4}$ & \hspace{-3mm}$\unit{N/m}$  \\
        EE-VI stiffness &   $6.91$ & $\num{1000}$ & \hspace{-3mm}$\unit{N/m}$ & $8.00$ & $\num{2993.7}$ & \hspace{-3mm}$\unit{N/m}$   \\
        \hline
        EE-ref damping &    $6.91$ & $\num{1000}$ & \hspace{-3mm}$\unit{Ns/m}$ & $8.01$ & $\num{30.1}$ & \hspace{-3mm}$\unit{Ns/m}$ \\
        Base-VI damping &   $6.91$ & $\num{1000}$ & \hspace{-3mm}$\unit{Ns/m}$ & $7.86$ & $\num{25.8}$ & \hspace{-3mm}$\unit{Ns/m}$  \\
        EE-VI damping &     $6.91$ & $\num{1000}$ & \hspace{-3mm}$\unit{Ns/m}$ & $7.97$ & $\num{28.8}$ & \hspace{-3mm}$\unit{Ns/m}$  \\
        J1 damping &        $6.91$ & $\num{1000}$ & \hspace{-3mm}$\unit{Ns/m}$ & $3.74$ & $\num{0.421}$ & \hspace{-3mm}$\unit{Ns/m}$   \\
        J4 damping &        $6.91$ & $\num{1000}$ & \hspace{-3mm}$\unit{Ns/m}$ & $6.27$ & $\num{5.29}$ & \hspace{-3mm}$\unit{Ns/m}$ \\
        \hline
    \end{tabular}
\end{table}

For the performance input $\w$, we consider several channels.
Disturbance forces are applied at two locations: a force $\d_{ee}$ is applied at the end effector and a second force $\d_{rcm}$ is applied to the instrument at the point closest to the remote centre of motion (RCM). We consider
noise $\n_{q}$ and $\n_{\dot{q}}$ applied to joint-level measurements of $\q$ and $\dot \q$, resulting in a $14$ dimensional noise vector (for the 7DOF robot). Finally, 
the surgeon inputs are taken into account through the reference signal $\bm{r}$, which
provides the desired end-effector displacement. Specifically, the end-effector reference 
$\bm{r}_{ee} = \bm{r}_0 + \bm{r}$, where $\bm{r}_0$ is the initial end-effector position.
We consider
\begin{equation}
    \label{eq:weighting_Ww}
        \begin{bmatrix}
		\d_{ee}  \\
            \d_{rcm} \\
            \n_q \\
		\n_{\dot q} \\
            \r
	\end{bmatrix} =
        \underbrace{\begin{bmatrix}
		W_{\d_{ee}} & 0 & 0 & 0 & 0 \\
            0 & W_{\d_{rcm}} & 0 & 0 & 0 \\
            0 & 0 & W_{\n_q} & 0 & 0 \\
		0 & 0 & 0 & W_{\n_{\dot q}} & 0 \\
            0 & 0 & 0 & 0 & W_{\r}
	\end{bmatrix}}_{W_w} \w  \,,      
\end{equation} 
where
$W_{\d_{ee}} = 8I$,
$W_{\d_{rcm}} = 4 I$,
$W_{\n_q} = 0.002I$,
$W_{\n_{\dot q}} = 0.1I$, and 
$W_{\r} =0.05I$.
We assume that the input $\w$ is a vector of sinusoids with maximal norm of $1$, 
therefore the disturbance forces peak at $8$N on the end-effector or $4$N  at the RCM, and the reference requires a displacement not larger than $5$ cm.
In this tuning setup, the weights are chosen to normalise quantities of different units and magnitudes, ensuring that reference motions, disturbances and noise signals are of a reasonable scale.
Noise on the velocity variables is greater than on the position variables, as
velocities are ordinarily obtained by numerical differentiation,
which amplifies noise.

\begin{figure}[t]
    \centering
    \includegraphics[scale=0.25]{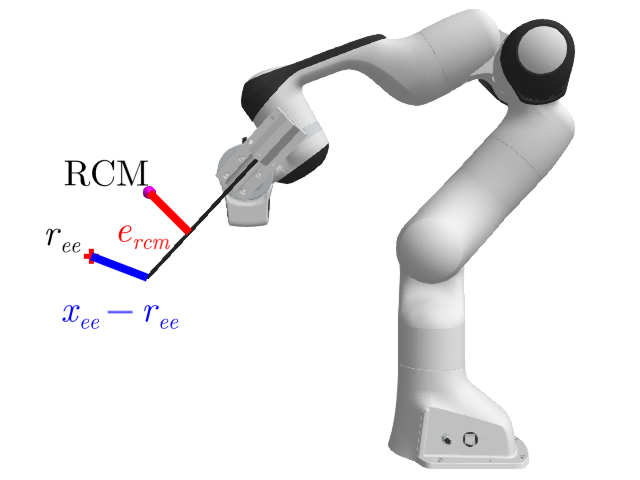}
    \caption{Performance outputs considered for the optimization of the parameters of the virtual instrument. $e_{rcm}$ is a length 2 vector representing the x-y position of the RCM in the instrument frame. The instrument axis is along the z axis of the instrument frame, thus $\parallel e_{rcm} \parallel_2$ is the shortest distance from the instrument axis to the RCM. 
    $x_{ee} - r_{ee}$ is the vector from the reference to the end-effector, and thus $\parallel x_{ee} - r_{ee} \parallel_2$ is the distance from the end-effector to the reference.
    }
    \label{fig:_lap_performance_output}
\end{figure}

The performance output $\y$ captures relevant position errors
\begin{equation}
        \y = Q
        \begin{bmatrix}
		\x_{ee} - \r_{ee} \\
        \e_{rcm} \\
	\end{bmatrix}       ,
\end{equation} 
where $\x_{ee}$ denotes the position of the end effector and 
$\e_{rcm}$ represents the `RCM error', that is, the minimal displacement
between the RCM and the longitudinal axis passing through the instrument, as shown in Fig. \ref{fig:_lap_performance_output}.
The output weighting matrix is chosen to emphasize performance at the end-effector, for which precision is more important.
\begin{equation}
    Q^2 = \begin{bmatrix}
        I_{3\times3} & 0 \\
        0 & 0.2I_{2\times2}
    \end{bmatrix}
\end{equation}
By minimizing the ratio between $\parallel\!\w\!\parallel_2$ and $\parallel\!\y\!\parallel_2$ we reduce the effect of noise, external disturbances, and reference signals on these position errors, improving accuracy. 

For the numerical optimization, we initialize the robot simulations at a rest position, with the end effector near the desired position and the instrument passing through the RCM.
We optimise over $7$ scenarios, summarized in Table \ref{tab:scenarios}. 
Each scenario includes a combination of noise $\n = [\n_q^T, \n_{\dot q}^T ]^T$, disturbance $\d = [\d_{ee}^T, \d_{rcm}^T ]^T$, and reference signal
$\r$, which are otherwise zero. We consider
\begin{subequations}
  \begin{align}
	\n_q &= \n_{\dot q} =  [1 \, 1\, 1\, 1\, 1\, 1\, 1]^T \frac{\sin(50t)}{\sqrt{7}}\\
	\d_{ee} &= \d_{rcm} = [1\,1\,1]^T \frac{\sin(2t)}{\sqrt{3}}\\
	\r &= [1\,0\,0]^T \frac{\sin(3t)}{\sqrt{3}}.
\end{align}  
\end{subequations}
Noise signals act at high frequency. 
Tracking is probed against slow disturbances and with references at $3 \,\mathrm{rad/s}$, under the assumption that this is the fastest frequency demanded by the surgeon. Good tracking performance at this frequency ensures good tracking at lower frequencies. 

As indicated by the cost function \eqref{eq:performance metric}, we optimize \emph{one} set of parameters by minimising the \emph{max} cost over all seven scenarios simultaneously, resulting in a single controller that gives the ``best-worst'' performance over the seven scenarios.

\begin{figure}[t]
    \centering
    \includegraphics{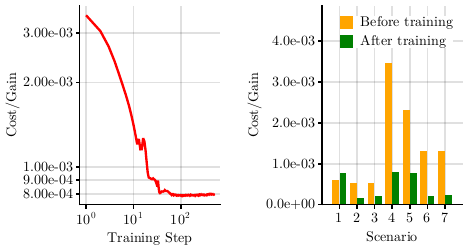}
    \caption{Training results. [Left] Log-log cost vs training step. [Right] Initial and final costs, $\frac{\parallel\y\parallel_2}{\parallel\w\parallel_2}$, on each individual scenario.}
    \label{fig:SurgeryBarChart}
\end{figure}
\begin{table}[t]
    \caption{Exogenous Signals present in each Scenario}
    \label{tab:scenarios}
	\centering
	\begin{tabular}{|l|lll|l|}
            \hline
                   &$\n$&$\d$&$\r$\\ \hline 
		Scenario 1 & x  &    &    \\ 
		Scenario 2 &    & x  &    \\
		Scenario 3 & x  & x  &    \\
		Scenario 4 &    &    & x  \\
		Scenario 5 & x  &    & x  \\
		Scenario 6 &    & x  & x  \\
		Scenario 7 & x  & x  & x  \\ \hline
	\end{tabular}
    \vspace{1pt}
\end{table}

\begin{figure}[t]
    \centering
    \includegraphics[]{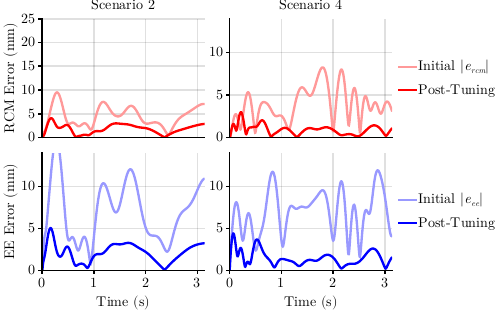}
    \caption{Comparison before and after optimization for Scenarios 2 and 4.}
    \label{fig:lap_performance_simulations}
\end{figure}

We test the optimized parameters on each scenario, as illustrated in Fig. \ref{fig:SurgeryBarChart}, right. 
The exogenous inputs included in each scenario are indicated in Table \ref{tab:scenarios}, alongside the (rounded) tuned parameters.
The comparison between the initial parameters and optimal tuning shows a consistent improvements in the cost metric in response to disturbances and references.
The case scenario 1 shows how the optimizing over all scenarios may increase the gain in a specific scenario. 
Plots of the achieved performance for Scenarios 2 and 4 are
shown in Fig. \ref{fig:lap_performance_simulations}. 
These show the resulting decrease in error at the end-effector and at the RCM in response to the pure disturbance and pure reference signals respectively.

\subsection{Scalability of the Optimal Laparoscopic Controller}
The small set of selected scenarios in the previous section allows for a detailed comparison of the achieved performance, summarized by the bar chart plot of Figure \ref{fig:SurgeryBarChart}. 
These examples are chosen for clarity, illustrating the method on a few representative signals that can be easily explained and compared. 
In this section, we demonstrate that the tuning technique generalises to a larger number of scenarios and more complex signals.

In what follows, scenarios are generated by randomly by sampling RCM positions, noise signals, disturbance signals, and reference signals. For each scenario, the RCM position is drawn from a Gaussian distribution centred on the nominal position (as in the previous Section), with a standard-deviation of $\SI{0.05}{m}$.
Each noise, disturbance, and reference signal is constructed as a sum of three sinusoids in a random direction of the form $\hat{n}_u \sum_{i=1}^{3} A_i \sin(\omega_i t)$. The frequencies $\omega_i$ are independently sampled from uniform distributions within the ranges shown in Table~\ref{tab:signal_ranges}.
The amplitudes $A_i$ are sampled in the range $[0, 1.0]$ and scaled by $W_w$, as defined in~\eqref{eq:weighting_Ww}.
The direction vector $\hat{n}_u$ is obtained by sampling the appropriate number of values from a unit normal distribution, assembling them into a vector, and normalizing.

Each scenario begins with the robot passing through the RCM, and the target tip position initially offset by $\SI{100}{mm}$ in the $x$-direction and $\SI{-100}{mm}$ in the $z$-direction.
By varying the RCM position and thus the robot pose, a wider portion of the system’s nonlinear dynamics is explored.
The randomised directions, amplitudes, and frequencies of the noise, disturbance, and reference signals further enrich the excitation space.
This demonstrates the flexibility of the approach: there is nothing special about the chosen class of signals. Any desired input signals, representative of the conditions under which performance is to be optimised, can be used.

Fifty scenarios are generated and jointly optimised.
Computation scales linearly with the number of scenarios, as each is simulated at every optimisation step. That is, in comparison to Section \ref{sec:SurgeryExample},
this extended scenario set requires about $50/7$ times more computation per step ($\num{3.93}\unit{s}$ vs.\ $\num{0.55}\unit{s}$ on a standard laptop).
The cost also increases with the number of tuned parameters due to forward-mode differentiation.

\begin{figure*}[t]
    \centering
    \includegraphics{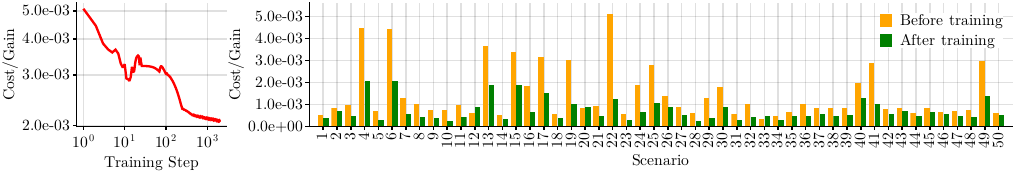}
    \caption{Training results on the larger set of scenarios [Left] Log-log cost vs training step. [Right] Initial and final costs, $\frac{\parallel\y\parallel_2}{\parallel\w\parallel_2}$, on each individual scenario.}
    \label{fig:extended_cost}
\end{figure*}

The results in Fig.~\ref{fig:extended_cost} and final parameters in Table~\ref{tab:extended_parameters} show a higher overall cost, reflecting the increased challenge of the randomised signals.
However, the optimised parameters follow similar trends—high stiffness and damping near the regularisation limits, with the notable exception of the EE-VI stiffness being much lower.

We remark that the goal of this section is to demonstrate the scalability of the propose optimization procedure, rather than to produce a 
complete control design for laparoscopic surgery.
The latter is beyond the scope of the current paper and would require a high-fidelity system model and carefully constructed excitation signals, representative of key operating conditions.

\begin{table}[t]
    \centering
    \caption{Frequency ranges for the extended set of excitation signals.}
	\label{tab:signal_ranges}
    \begin{tabular}{|l?{1.5pt}cc|}
        \hline
        & \multicolumn{2}{c|}{Freq. range (rad/s)} \\ 
        Signal & Min & Max \\
        \hline
        $\n_q$              & $25.0$ & $100.0$ \\
        $\n_{\dot{q}}$      & $25.0$ & $100.0$  \\
        $\d_{ee}$           & $0.0$ & $5.0$ \\
        $\d_{rcm}$          & $0.0$ & $5.0$\\
        $\r$                & $0.0$ & $5.0$ \\
        \hline
    \end{tabular}
\end{table}

% \begin{figure}[htbp]
%     \centering
%     \includegraphics{figures/FullSurgeryOptimization.pdf}
%     \caption{Surgery optimization figure.}
%     \label{fig:SurgeryOptimization}
% \end{figure}

% Describe cost function
% Noise, disturbance, desired magnitudes and gains

\section{Experimental validation}
\label{sec:ExperimentalValidation}

\begin{figure*}[htbp]
    \begin{tabular}{cccccc}
        
          \includegraphics[width=0.166\textwidth]{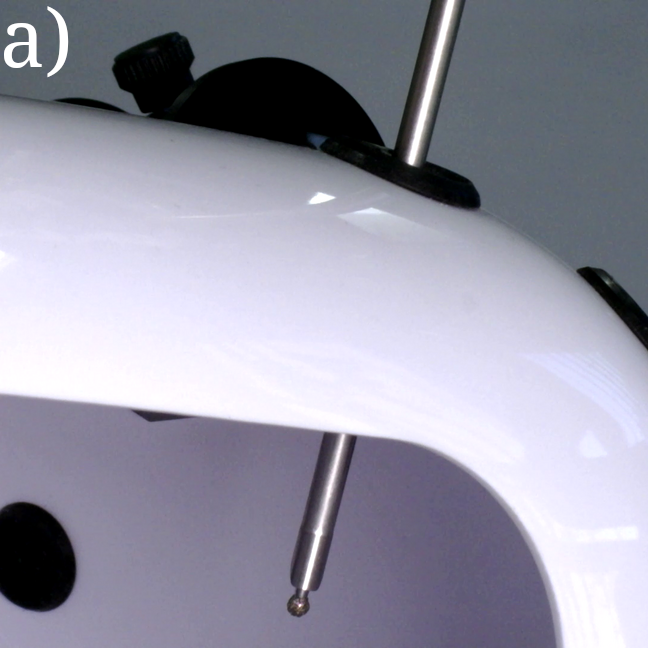}
        & \includegraphics[width=0.166\textwidth]{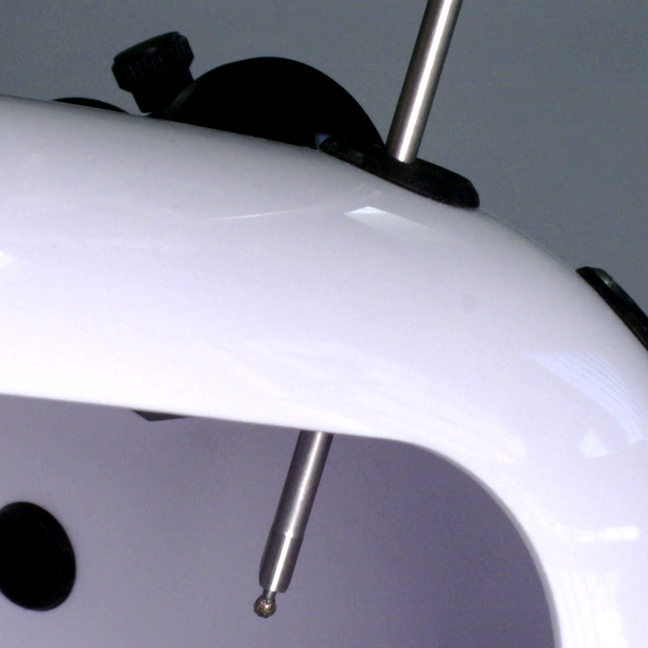}
        & \includegraphics[width=0.166\textwidth]{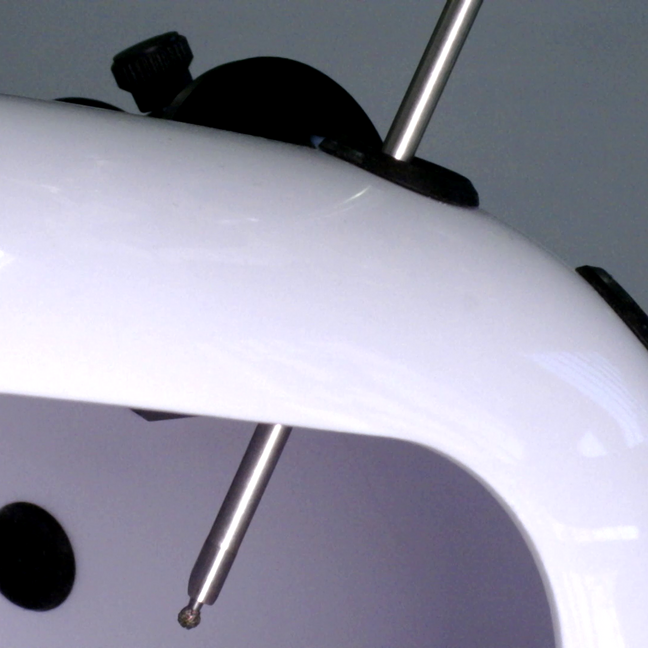}
        & \includegraphics[width=0.166\textwidth]{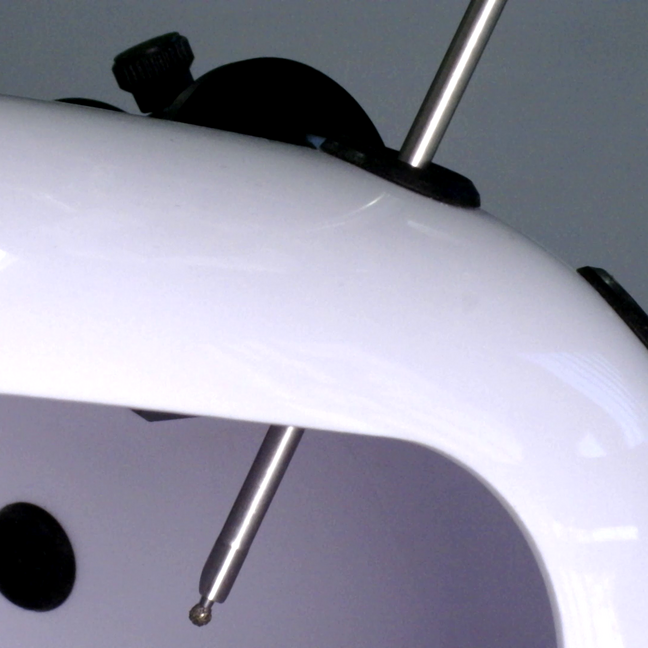}
        & \includegraphics[width=0.166\textwidth]{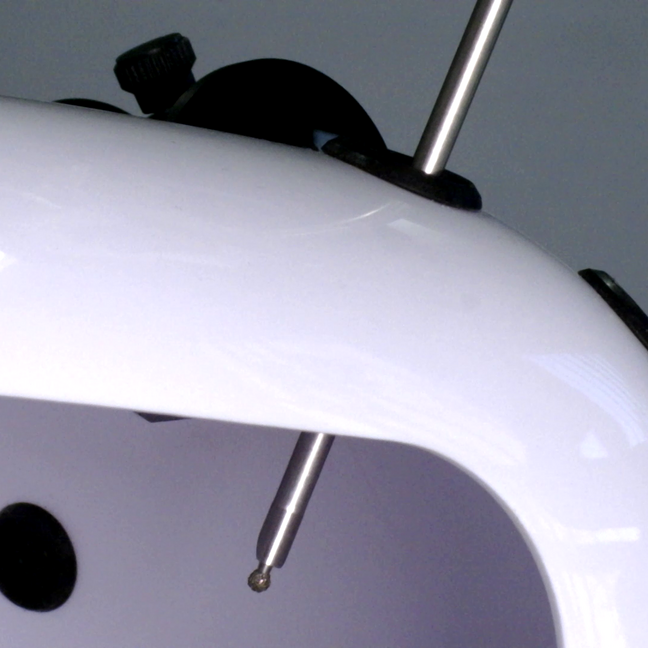}
        & \includegraphics[width=0.166\textwidth]{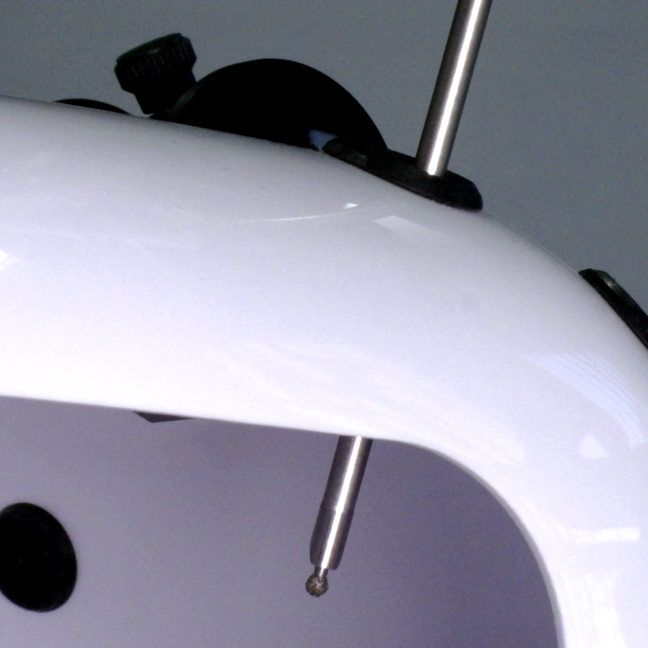}
        \\
          \includegraphics[width=0.166\textwidth]{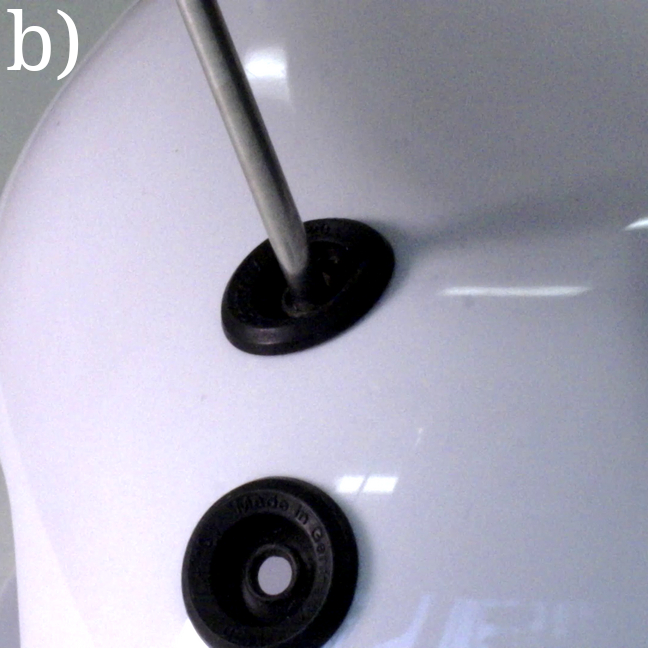}
        & \includegraphics[width=0.166\textwidth]{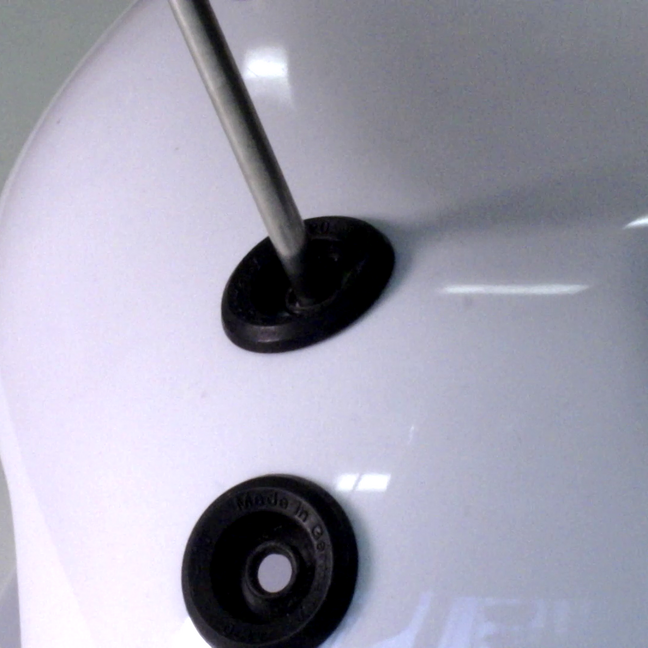}
        & \includegraphics[width=0.166\textwidth]{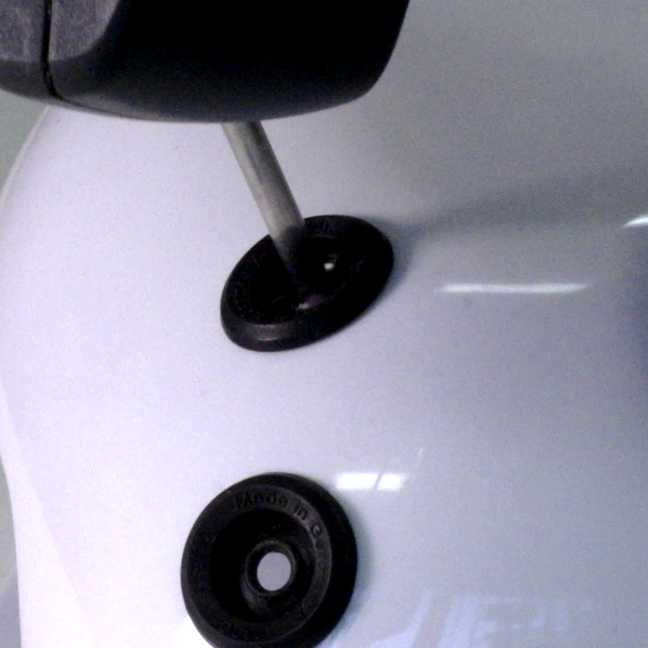}
        & \includegraphics[width=0.166\textwidth]{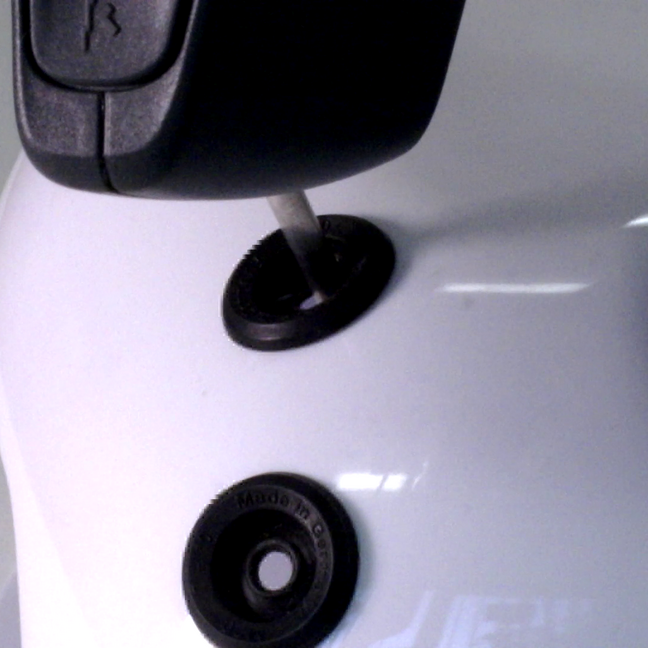}
        & \includegraphics[width=0.166\textwidth]{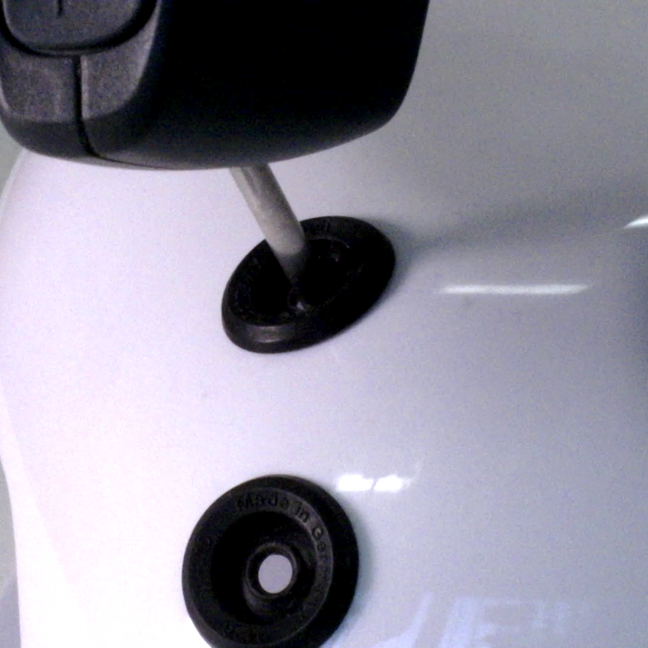}
        & \includegraphics[width=0.166\textwidth]{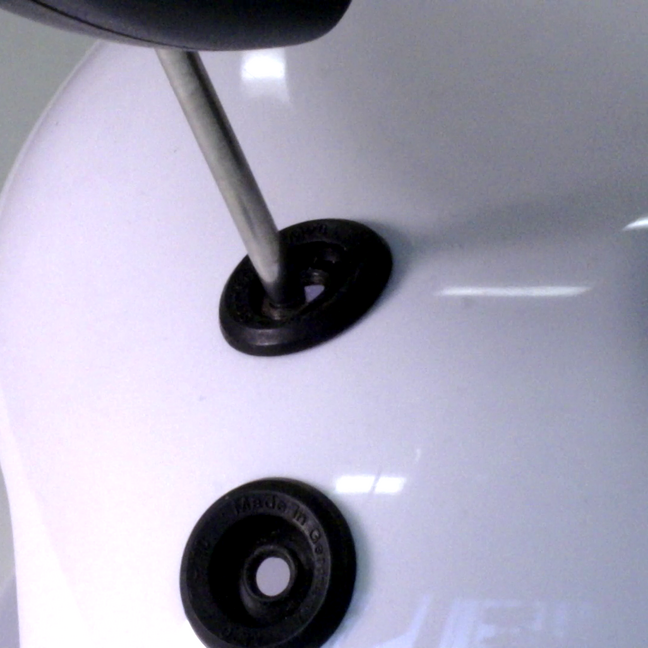}
    \end{tabular}
           
    \centering   
    \includegraphics[height=150px]{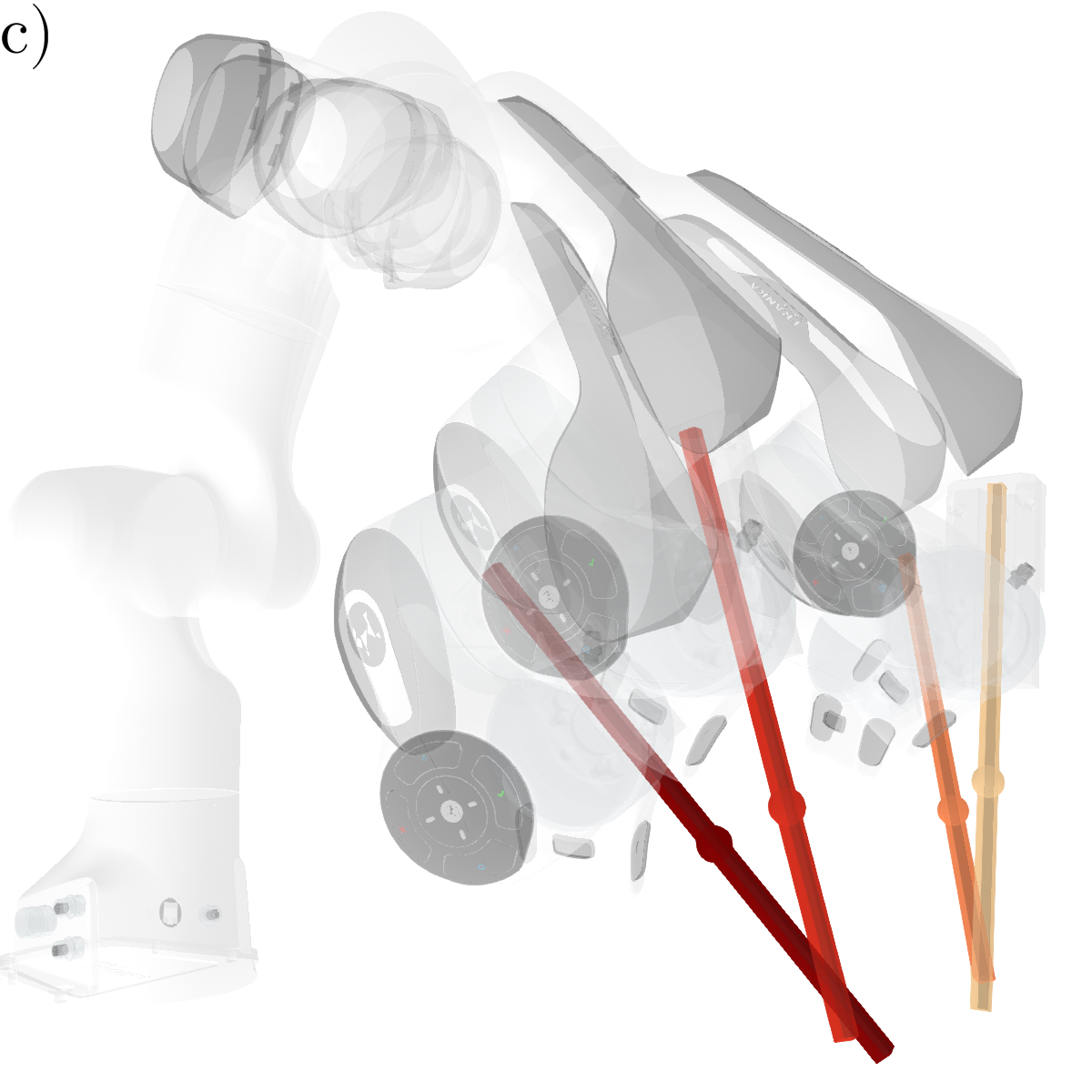}
    \includegraphics[height=150px]{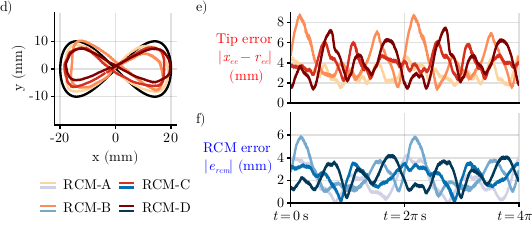}
    \caption{Experimental performance of the controller while passing through 4 different 'incisions' on a laparoscopy training phantom, labelled RCM-A through D. Video frames showing RCM-A at t=0s, 1s, \dots 5s, from the side A) and top B) of the phantom respectively. C) Initial poses for of the robot/virtual instrument. D) Tip error x-y trajectories. E) Tip error magnitude. F) RCM error magnitudes. All positions are computed using the kinematics of the robot and measured joint angles.}
    \label{fig:lap_performance_experiment}
\end{figure*}

We validate the optimal controller of Section \ref{sec:SurgeryExample} experimentally on a Franka Emika manipulator. The robot is illustrated in Fig. \ref{fig:VirtualInstrument}, and it is tested passing through 4 different `incisions` on the surgical phantom of Fig. \ref{fig:RMIS}, which are openings in the plastic body with rubber aperture.
The RCM location is determined by moving the robot so that the tip of the instrument is at the centre of the aperture, and measuring the location.
Then the robot is moved by hand so that the instrument is inserted into the phantom, and the controller is started.
To mimic a surgical reference, we apply a reference signal around the starting tip location of $\bm{r}=[0.02 \sin(t), 0.01\sin(2t), 0]^T$, a figure-8 motion in the x-y plane with peak-peak amplitude 40mm, with frequency components at $\SI{1}{rad/s}$ and $\SI{2}{rad/s}$ as shown in Fig. \ref{fig:lap_performance_experiment}(d).

The controller runs on a PC that communicates with the robot via a C++ program based on the ``libfranka" library.
The virtual model controller is written in Julia using the \emph{VMRobotControl.jl} library \cite{VMRobotControl}, developed by the authors, to simulate the virtual instrument, and compute the desired robot torques due to the interface components.
This library is used both for the tuning and implementation of the controller. During the simulation stage, the controller is interfaced with a simulated Franka-Emika robot. At implementation stage, the simulated robot is replaced by
the real robot, via rerouting of the communication.
The virtual model controller runs on a linux PC with the real-time kernel patch and transmits the desired joint torques to the robot at a rate of $\SI{1}{kHz}$, which are then realized by the robot's internal joint torque controllers.

\begin{table}[t]
    \centering
    \caption{Extended scenario tuned parameters (rounded).}
	\label{tab:extended_parameters}
    \begin{tabular}{|l?{1.5pt}lrl|}
        \hline
        & \multicolumn{3}{c|}{Final} \\ 
        Description & $\,\,\,\,{\bm{\theta}}$ & \qquad$\hat{\bm{\theta}}$ &  \\
        \hline
        EE-ref stiffness &  $8.02$ & $\num{3040.2}$ & \hspace{-3mm}$\unit{N/m}$   \\
        Base-VI stiffness & $8.01$ & $\num{3002.4}$ & \hspace{-3mm}$\unit{N/m}$  \\
        EE-VI stiffness &   $5.53$ & $\num{252.7}$ & \hspace{-3mm}$\unit{N/m}$   \\
        \hline
        EE-ref damping &    $8.01$ & $\num{30.1}$ & \hspace{-3mm}$\unit{Ns/m}$ \\
        Base-VI damping &   $7.94$ & $\num{28.1}$ & \hspace{-3mm}$\unit{Ns/m}$  \\
        EE-VI damping &     $8.02$ & $\num{30.4}$ & \hspace{-3mm}$\unit{Ns/m}$  \\
        J1 damping &        $7.68$ & $\num{21.6}$ & \hspace{-3mm}$\unit{Ns/m}$   \\
        J4 damping &        $5.84$ & $\num{3.44}$ & \hspace{-3mm}$\unit{Ns/m}$ \\
        \hline
    \end{tabular}
\end{table}

The experimental errors of the tip and RCM shown in Fig. \ref{fig:lap_performance_experiment}(e) and Fig. \ref{fig:lap_performance_experiment}(f) vary depending on pose but generally remain well below 10 mm. Their statistics are summarised in Table \ref{tab:rcm_tip_errors}. Although this performance is not accurate enough in terms of absolute placement of the surgical tool tip, we should recall that the tool reference is governed by a surgeon who can adapt the reference accordingly to visual feedback and insights from experience. In that sense, we would argue that an interpretable and compliant response of the robot is more important than absolute accuracy. In general, reference tracking can be improved by increasing the stiffness of the virtual mechanism, if the bandwidth of the actuators allows for it (this is a limitation of the nonsurgical manipulator used in the experiments). Alternatively, performance can be improved by introducing integral action within a slow outer feedback loop, as shown in \cite{Larby2025},
or by considering friction correction terms, if a precise model can be estimated. These additional control terms
do not necessarily preserve passivity but are entirely compatible with the virtual mechanism approach.

Because of our approach and the structure of the controller, particularly its passivity, we expect that the interconnection of the virtual model controller and the real robot should remain stable.
This is despite mismatches between the robot model used for the tuning and the robot itself, such as uncertainty in the inertia parameters and the lack of any joint-friction model.
Due to the constraint of passing through the {\color{blue}rubber aperture} of the surgical phantom, all of the 4 poses tested,
represented in Fig. \ref{fig:lap_performance_experiment}(c), deviate from the training pose. This results in a different behaviour to that experienced in simulation and tuning. However, due to the structure of the controller, we still expect reasonable performances.

Notably, if the sensing-communication-actuation loop accurately renders the desired torques, the controller's passivity guarantees that the closed-loop system will be stable. This stability holds even when the real robot's dynamics do not match the simulated dynamics and under interaction with external passive systems, such as the instrument interacting with the phantom.
However, a loss of performance is expected due the mismatch between ideal and real parameters
of the robot.
 
\setlength{\tabcolsep}{0pt}
\renewcommand{\arraystretch}{0} 

\setlength{\tabcolsep}{6pt}
\renewcommand{\arraystretch}{1} 

\begin{table}[t]
\centering
\caption{RCM and Tip errors mean $\pm$ standard deviation.}
\label{tab:rcm_tip_errors}
\begin{tabular}{|c|c|c|}
\hline
\textbf{Test} & \textbf{RCM Error (mm)} & \textbf{Tip Error (mm)} \\
\hline
RCM-A & 1.92 $\pm$ 1.10 & 4.52 $\pm$ 3.77 \\
RCM-B & 2.34 $\pm$ 1.43 & 5.70 $\pm$ 3.69 \\
RCM-C & 2.27 $\pm$ 0.79 & 5.39 $\pm$ 3.65 \\
RCM-D & 2.15 $\pm$ 0.99 & 5.30 $\pm$ 3.65 \\
\hline
\end{tabular}
\end{table}

\section{Conclusions} \label{sec:Conclusions}

We have presented an approach that combines the strengths of virtual-mechanisms with iterative tuning via optimization using automatic differentiation.
This approach incorporates nonlinear dynamics, nonlinear controllers and nonlinear cost functions, allowing it to be flexible and general.
The approach takes advantage of the chosen control structures to ensure structural properties of the controller, such as passivity.
Passivity guarantees closed-loop stability, both in simulations and experiments. This facilitates iterative tuning in simulations, and helps with the sim-to-real transition, despite model uncertainties.

Our approach has two fundamental limitations. One is intrinsic to the non-convexity of the optimization problem \eqref{eq:performance metric}, leading to local minima and lack of convergence guarantees. However, for the parameterization and cost functions of our examples, we have shown that the optimization converges, recovering classical linear results (Section \ref{sec:PDcart}), 
and handling the complexity of the Franka-Emika manipulator (Section \ref{sec:SurgeryExample}).
The other limitation is the dependence of performance optimization on the accuracy of the robot model, which is challenging, particularly with phenomena such as non-linear friction dependent on load / temperature \cite{Bittencourt2012a} and transmission non-linearities \cite{Kennedy2003, Ma2018}.

This virtual mechanism approach to passivity-based control shifts design of robot controllers and their parameters into a space familiar to engineers.
The design of the controller structure is framed in the language of mechanisms: components, springs, dampers, joints, and links.
Within the framework of virtual mechanism controllers there is much to explore: such the multitude of possible components, including the examples of nonlinear springs and dampers given here.

Optimal virtual model control could be used to help design structures linking leader/follower manipulators, with an adaptation of the cost function to optimise for teleoperative transparency.
Alternatively, the automatic differentiation approach presented here could be used to optimise the performance in model-mediated teleoperation \cite{Xu2016}.

In the limit of highly parameterized and highly numerous components, our approach looks similar to physics-informed neural networks, as seen in \cite{Massaroli2022} where the controller is a passive highly parameterized neural network. The trade-off between the flexibility of the controller for performance optimization and the physical intuition provided by virtual mechanisms can be explored by placing highly parameterized components at specific points on the virtual mechanism. Furthermore, the adoption of highly parameterized structural parts can further improve performance while retaining the physical intuition on the behavior of the controller / virtual mechanism. Future research will focus on these important aspects. We will also look at data-driven tuning to reduce the gap between simulation and experimental performances. 

\bibliographystyle{IEEEtran}
\bibliography{library, extrabib}

@online{Larby2025,
  title = {Collaborative {{Drill Alignment}} in {{Surgical Robotics}}},
  author = {Larby, Daniel and Kershaw, Joshua and Allen, Matthew and Forni, Fulvio},
  date = {2025-02-28},
  eprint = {2503.05791},
  eprinttype = {arXiv},
  eprintclass = {eess},
  doi = {10.48550/arXiv.2503.05791},
  note = {{{arXiv}}:2503.05791 [eess.SY]},
  pubstate = {prepublished}
}

@article{Xu2016,
  title = {Model-{{Mediated Teleoperation}}: {{Toward Stable}} and {{Transparent Teleoperation Systems}}},
  shorttitle = {Model-{{Mediated Teleoperation}}},
  author = {Xu, Xiao and Cizmeci, Burak and Schuwerk, Clemens and Steinbach, Eckehard},
  journal={IEEE Access},
  year={2016},
  volume={4},
  pages={425-449},
  url={https://api.semanticscholar.org/CorpusID:9121437},
  doi = {10.1109/ACCESS.2016.2517926},
}

@inproceedings{Ekkelenkamp2007,
  title = {Evaluation of a {{Virtual Model Control}} for the Selective Support of Gait Functions Using an Exoskeleton},
  booktitle = {2007 {{IEEE}} 10th {{International Conference}} on {{Rehabilitation Robotics}}},
  author = {Ekkelenkamp, R. and Veltink, Peter and Stramigioli, Stefano and {van der Kooij}, H},
  year = {2007},
  month = jun,
  pages = {693--699},
  issn = {1945-7901},
  doi = {10.1109/ICORR.2007.4428501},
  urldate = {2024-12-09}
}

@misc{VMRobotControl,
  title = {{{VMRobotControl}}.Jl},
  author = {Larby, Daniel},
  date = {2024},
  url = {https://github.com/Cambridge-Control-Lab/VMRobotControl.jl}
}

@inproceedings{Averta2020,
  title = {Enhancing {{Robot-Environment Physical Interaction}} via {{Optimal Impedance Profiles}}},
  booktitle = {2020 8th {{IEEE RAS}}/{{EMBS International Conference}} for {{Biomedical Robotics}} and {{Biomechatronics}} ({{BioRob}})},
  author = {Averta, Giuseppe and Hogan, Neville},
  year = {2020},
  month = nov,
  pages = {973--980},
  issn = {2155-1782},
  doi = {10.1109/BioRob49111.2020.9224382}
}

@article{Bittencourt2012a,
  title = {Static {{Friction}} in a {{Robot Joint}}---{{Modeling}} and {{Identification}} of {{Load}} and {{Temperature Effects}}},
  author = {Bittencourt, Andr{\'e} Carvalho and Gunnarsson, Svante},
  year = {2012},
  month = sep,
  journal = {Journal of Dynamic Systems, Measurement, and Control},
  volume = {134},
  number = {5},
  pages = {051013},
  issn = {0022-0434, 1528-9028},
  doi = {10.1115/1.4006589},
  urldate = {2021-09-13},
  langid = {english}
}

@article{Bloch2000,
  title = {Controlled {{Lagrangians}} and the Stabilization of Mechanical Systems. {{I}}. {{The}} First Matching Theorem},
  author = {Bloch, A.M. and Leonard, N.E. and Marsden, J.E.},
  year = {2000},
  month = dec,
  journal = {IEEE Transactions on Automatic Control},
  volume = {45},
  number = {12},
  pages = {2253--2270},
  issn = {1558-2523},
  doi = {10.1109/9.895562},
  urldate = {2024-09-03}
}

@article{Bowyer2014,
  title = {Active {{Constraints}}/{{Virtual Fixtures}}: {{A Survey}}},
  author = {Bowyer, Stuart A and Davies, Brian L and {Rodriguez y Baena}, Ferdinando},
  year = {2014},
  month = feb,
  journal = {IEEE Transactions on Robotics},
  volume = {30},
  number = {1},
  pages = {138--157},
  publisher = {IEEE},
  issn = {1552-3098},
  doi = {10.1109/TRO.2013.2283410}
}

@book{Boyd1994,
  title = {Linear {{Matrix Inequalities}} in {{System}} and {{Control Theory}}},
  author = {Boyd, Stephen and El Ghaoui, Laurent and Feron, Eric and Balakrishnan, Venkataramanan},
  year = {1994},
  month = jan,
  publisher = {{Society for Industrial and Applied Mathematics}},
  doi = {10.1137/1.9781611970777},
  urldate = {2022-03-27},
  isbn = {978-0-89871-485-2 978-1-61197-077-7},
  langid = {english}
}

@book{Brunton2019,
  title = {Data-{{Driven Science}} and {{Engineering}}: {{Machine Learning}}, {{Dynamical Systems}}, and {{Control}}},
  shorttitle = {Data-{{Driven Science}} and {{Engineering}}},
  author = {Brunton, Steven L. and Kutz, J. Nathan},
  year = {2019},
  publisher = {Cambridge University Press},
  address = {Cambridge},
  doi = {10.1017/9781108380690},
  urldate = {2024-09-03}
}

@article{Chen2019,
  title = {Contact Force Control and Vibration Suppression in Robotic Polishing with a Smart End Effector},
  author = {Chen, Fan and Zhao, Huan and Li, Dingwei and Chen, Lin and Tan, Chao and Ding, Han},
  year = {2019},
  month = jun,
  journal = {Robotics and Computer-Integrated Manufacturing},
  volume = {57},
  pages = {391--403},
  issn = {0736-5845},
  doi = {10.1016/j.rcim.2018.12.019},
  urldate = {2023-04-19},
  langid = {english}
}

@article{Chen2020a,
  title = {Virtual {{Model Control}} for {{Quadruped Robots}}},
  author = {Chen, Guangrong and Guo, Sheng and Hou, Bowen and Wang, Junzheng},
  year = {2020},
  journal = {IEEE Access},
  volume = {8},
  pages = {140736--140751},
  issn = {2169-3536},
  doi = {10.1109/ACCESS.2020.3013434}
}

@article{Chopra2022,
  title = {Passivity-{{Based Control}} of {{Robots}}: {{Theory}} and {{Examples}} from the {{Literature}}},
  shorttitle = {Passivity-{{Based Control}} of {{Robots}}},
  author = {Chopra, Nikhil and Fujita, Masayuki and Ortega, Romeo and Spong, Mark W.},
  year = {2022},
  month = apr,
  journal = {IEEE Control Systems Magazine},
  volume = {42},
  number = {2},
  pages = {63--73},
  issn = {1941-000X},
  doi = {10.1109/MCS.2021.3139722}
}

@inproceedings{Desai2014,
  title = {Virtual Model Control for Dynamic Lateral Balance},
  booktitle = {2014 {{IEEE-RAS International Conference}} on {{Humanoid Robots}}},
  author = {Desai, Ruta and Geyer, Hartmut and Hodgins, Jessica K.},
  year = {2014},
  month = nov,
  pages = {856--861},
  publisher = {IEEE},
  address = {Madrid, Spain},
  doi = {10.1109/HUMANOIDS.2014.7041464},
  urldate = {2022-10-04},
  isbn = {978-1-4799-7174-9},
  langid = {english}
}

@book{Featherstone2008,
  title = {Rigid {{Body Dynamics Algorithms}}},
  author = {Featherstone, Roy},
  year = {2008},
  publisher = {Springer US},
  address = {Boston, MA},
  doi = {10.1007/978-1-4899-7560-7},
  urldate = {2022-05-04},
  isbn = {978-0-387-74314-1},
  langid = {english}
}

@article{Funda1996,
  title = {Constrained {{Cartesian}} Motion Control for Teleoperated Surgical Robots},
  author = {Funda, Janez and Taylor, R.H. and Eldridge, Benjamin and Gomory, Stephen and Gruben, K.G.},
  year = {1996},
  month = jun,
  journal = {IEEE Transactions on Robotics and Automation},
  volume = {12},
  number = {3},
  issn = {1042296X},
  doi = {10.1109/70.499826}
}

@article{Gaz2019,
  title = {Dynamic {{Identification}} of the {{Franka Emika Panda Robot With Retrieval}} of {{Feasible Parameters Using Penalty-Based Optimization}}},
  author = {Gaz, Claudio and Cognetti, Marco and Oliva, Alexander and Robuffo Giordano, Paolo and De Luca, Alessandro},
  year = {2019},
  month = oct,
  journal = {IEEE Robotics and Automation Letters},
  volume = {4},
  number = {4},
  pages = {4147--4154},
  issn = {2377-3766, 2377-3774},
  doi = {10.1109/LRA.2019.2931248},
  urldate = {2022-11-28},
  langid = {english}
}

@article{Giftthaler2017,
  title = {Automatic {{Differentiation}} of {{Rigid Body Dynamics}} for {{Optimal Control}} and {{Estimation}}},
  author = {Giftthaler, Markus and Neunert, Michael and St{\"a}uble, Markus and Frigerio, Marco and Semini, Claudio and Buchli, Jonas},
  year = {2017},
  month = nov,
  journal = {Advanced Robotics},
  volume = {31},
  number = {22},
  eprint = {1709.03799},
  primaryclass = {cs},
  pages = {1225--1237},
  issn = {0169-1864, 1568-5535},
  doi = {10.1080/01691864.2017.1395361},
  urldate = {2024-01-23},
  archiveprefix = {arXiv}
}

@inproceedings{Giftthaler2018,
  title = {The {{Control Toolbox}} - {{An Open-Source C}}++ {{Library}} for {{Robotics}}, {{Optimal}} and {{Model Predictive Control}}},
  booktitle = {2018 {{IEEE International Conference}} on {{Simulation}}, {{Modeling}}, and {{Programming}} for {{Autonomous Robots}} ({{SIMPAR}})},
  author = {Giftthaler, Markus and Neunert, Michael and St{\"a}uble, Markus and Buchli, Jonas},
  year = {2018},
  month = may,
  eprint = {1801.04290},
  primaryclass = {cs, math},
  pages = {123--129},
  doi = {10.1109/SIMPAR.2018.8376281},
  urldate = {2022-11-15},
  archiveprefix = {arXiv}
}

@misc{Greydanus2019,
  title = {Hamiltonian {{Neural Networks}}},
  author = {Greydanus, Sam and Dzamba, Misko and Yosinski, Jason},
  year = {2019},
  month = sep,
  number = {arXiv:1906.01563},
  eprint = {1906.01563},
  primaryclass = {cs},
  publisher = {arXiv},
  doi = {10.48550/arXiv.1906.01563},
  urldate = {2023-08-08},
  archiveprefix = {arXiv}
}

@inproceedings{Hogan1984,
  title = {Impedance {{Control}}: {{An Approach}} to {{Manipulation}}},
  shorttitle = {Impedance {{Control}}},
  booktitle = {1984 {{American Control Conference}}},
  author = {Hogan, Neville},
  year = {1984},
  month = jun,
  pages = {304--313},
  doi = {10.23919/ACC.1984.4788393}
}

@article{Hogan2022,
  title = {Contact and {{Physical Interaction}}},
  author = {Hogan, Neville},
  year = {2022},
  month = may,
  journal = {Annual Review of Control, Robotics, and Autonomous Systems},
  volume = {5},
  number = {1},
  pages = {annurev-control-042920-010933},
  issn = {2573-5144, 2573-5144},
  doi = {10.1146/annurev-control-042920-010933},
  urldate = {2021-11-17},
  langid = {english}
}

@misc{Howell2023,
  title = {Dojo: {{A Differentiable Physics Engine}} for {{Robotics}}},
  shorttitle = {Dojo},
  author = {Howell, Taylor A. and Cleac'h, Simon Le and Br{\"u}digam, Jan and Kolter, J. Zico and Schwager, Mac and Manchester, Zachary},
  year = {2023},
  month = mar,
  number = {arXiv:2203.00806},
  eprint = {2203.00806},
  primaryclass = {cs},
  publisher = {arXiv},
  doi = {10.48550/arXiv.2203.00806},
  urldate = {2024-02-02},
  archiveprefix = {arXiv}
}

@article{Hsieh2021,
  title = {The {{Limits}} of {{Min-Max Optimization Algorithms}}: {{Convergence}} to {{Spurious Non-Critical Sets}}},
  author = {Hsieh, Ya-Ping and Mertikopoulos, Panayotis and Cevher, Volkan},
  year = {2021},
  journal = {International Conference on Machine Learning},
  volume = {38},
  issn = {139:4337-4348},
  langid = {english}
}

@inproceedings{JianjuenHrr1998,
  title = {Adaptive Virtual Model Control of a Bipedal Walking Robot},
  booktitle = {Proceedings. {{IEEE International Joint Symposia}} on {{Intelligence}} and {{Systems}} ({{Cat}}. {{No}}.{{98EX174}})},
  author = {{Jianjuen Hrr} and Pratt, J. and {Chee-Meng Chew} and Herr, H. and Pratt, G.},
  year = {1998},
  pages = {245--251},
  publisher = {IEEE Comput. Soc},
  address = {Rockville, MD, USA},
  doi = {10.1109/IJSIS.1998.685453},
  urldate = {2021-07-21},
  isbn = {978-0-8186-8548-4},
  langid = {english}
}

@inproceedings{Joly1995,
  title = {Imposing Motion Constraints to a Force Reflecting Telerobot through Real-Time Simulation of a Virtual Mechanism},
  booktitle = {Proceedings of 1995 {{IEEE International Conference}} on {{Robotics}} and {{Automation}}},
  author = {Joly, L.D. and Andriot, C.},
  year = {1995},
  month = may,
  volume = {1},
  pages = {357-362 vol.1},
  issn = {1050-4729},
  doi = {10.1109/ROBOT.1995.525310},
  urldate = {2024-07-11}
}

@misc{Kennedy2003,
  title = {Estimation and Modeling of the Harmonic Drive Transmission in the {{Mitsubishi PA-10}} Robot Arm},
  author = {Kennedy, Christopher W. and Desai, Jaydev P.},
  year = {2003}
}

@article{Khatib1987,
  title = {A Unified Approach for Motion and Force Control of Robot Manipulators: {{The}} Operational Space Formulation},
  author = {Khatib, Oussama},
  year = {1987},
  month = feb,
  journal = {IEEE Journal on Robotics and Automation},
  volume = {3},
  number = {1},
  pages = {43--53},
  issn = {0882-4967},
  doi = {10.1109/JRA.1987.1087068}
}

@article{Kim2021,
  title = {Stiff Neural Ordinary Differential Equations},
  author = {Kim, Suyong and Ji, Weiqi and Deng, Sili and Ma, Yingbo and Rackauckas, Christopher},
  year = {2021},
  month = sep,
  journal = {Chaos: An Interdisciplinary Journal of Nonlinear Science},
  volume = {31},
  number = {9},
  pages = {093122},
  issn = {1054-1500, 1089-7682},
  doi = {10.1063/5.0060697},
  urldate = {2023-11-22},
  langid = {english}
}

@article{Lachner2022,
  title = {Shaping {{Impedances}} to {{Comply With Constrained Task Dynamics}}},
  author = {Lachner, Johannes and Allmendinger, Felix and Stramigioli, Stefano and Hogan, Neville},
  year = {2022},
  month = oct,
  journal = {IEEE Transactions on Robotics},
  volume = {38},
  number = {5},
  pages = {2750--2767},
  issn = {1941-0468},
  doi = {10.1109/TRO.2022.3153949}
}

@misc{Larby2022,
  title = {A {{Generalized Approach}} to {{Impedance Control Design}} for {{Robotic Minimally Invasive Surgery}}},
  author = {Larby, Daniel and Forni, Fulvio},
  year = {2022},
  month = dec,
  number = {arXiv:2212.11244},
  eprint = {2212.11244},
  primaryclass = {cs, eess},
  publisher = {arXiv},
  doi = {10.48550/arXiv.2212.11244},
  urldate = {2023-02-20},
  archiveprefix = {arXiv}
}

@inproceedings{Larby2022b,
  title = {A {{Passivity Preserving H-infinity Synthesis Technique}} for {{Robot Control}}},
  booktitle = {2022 {{IEEE}} 61st {{Conference}} on {{Decision}} and {{Control}} ({{CDC}})},
  author = {Larby, Daniel and Forni, Fulvio},
  year = {2022},
  month = dec,
  pages = {1416--1422},
  issn = {2576-2370},
  doi = {10.1109/CDC51059.2022.9993347}
}

@inproceedings{Lawrence1988,
  title = {Impedance Control Stability Properties in Common Implementations},
  booktitle = {1988 {{IEEE International Conference}} on {{Robotics}} and {{Automation Proceedings}}},
  author = {Lawrence, D.A.},
  year = {1988},
  month = apr,
  pages = {1185-1190 vol.2},
  doi = {10.1109/ROBOT.1988.12222}
}

@inproceedings{Ma2018,
  title = {Investigation of the Friction Behavior of Harmonic Drive Gears at Low Speed Operation},
  booktitle = {2018 {{IEEE International Conference}} on {{Mechatronics}} and {{Automation}} ({{ICMA}})},
  author = {Ma, Donghui and Yan, Shaoze and Yin, Zhixiang and Yang, Yunqiang},
  year = {2018},
  month = aug,
  pages = {1382--1388},
  publisher = {IEEE},
  address = {Changchun},
  doi = {10.1109/ICMA.2018.8484324},
  urldate = {2024-11-05},
  isbn = {978-1-5386-6074-4 978-1-5386-6075-1}
}

@inproceedings{Ma2021,
  title = {A {{Comparison}} of {{Automatic Differentiation}} and {{Continuous Sensitivity Analysis}} for {{Derivatives}} of {{Differential Equation Solutions}}},
  booktitle = {2021 {{IEEE High Performance Extreme Computing Conference}} ({{HPEC}})},
  author = {Ma, Yingbo and Dixit, Vaibhav and Innes, Michael J and Guo, Xingjian and Rackauckas, Chris},
  year = {2021},
  month = sep,
  pages = {1--9},
  publisher = {IEEE},
  address = {Waltham, MA, USA},
  doi = {10.1109/HPEC49654.2021.9622796},
  urldate = {2023-11-22},
  isbn = {978-1-66542-369-4},
  langid = {english}
}

@inproceedings{Marinho2019,
  title = {A {{Unified Framework}} for the {{Teleoperation}} of {{Surgical Robots}} in {{Constrained Workspaces}}},
  booktitle = {2019 {{International Conference}} on {{Robotics}} and {{Automation}} ({{ICRA}})},
  author = {Marinho, Murilo M. and Adorno, Bruno V. and Harada, Kanako and Deie, Kyoichi and Deguet, Anton and Kazanzides, Peter and Taylor, Russell H. and Mitsuishi, Mamoru},
  year = {2019},
  month = may,
  pages = {2721--2727},
  publisher = {IEEE},
  address = {Montreal, QC, Canada},
  doi = {10.1109/ICRA.2019.8794363},
  urldate = {2023-10-10},
  isbn = {978-1-5386-6027-0},
  langid = {english}
}

@article{Massaroli2022,
  title = {Optimal {{Energy Shaping}} via {{Neural Approximators}}},
  author = {Massaroli, Stefano and Poli, Michael and Califano, Federico and Park, Jinkyoo and Yamashita, Atsushi and Asama, Hajime},
  year = {2022},
  month = sep,
  journal = {SIAM Journal on Applied Dynamical Systems},
  volume = {21},
  number = {3},
  pages = {2126--2147},
  publisher = {{Society for Industrial and Applied Mathematics}},
  doi = {10.1137/21M1414279},
  urldate = {2023-06-06}
}

@article{Ortega1989,
  title = {Adaptive Motion Control of Rigid Robots: {{A}} Tutorial},
  author = {Ortega, Romeo and Spong, Mark W.},
  year = {1989},
  journal = {Automatica},
  volume = {25},
  number = {6},
  pages = {877--888},
  issn = {00051098},
  doi = {10.1016/0005-1098(89)90054-X}
}

@book{Ortega1998,
  title = {Passivity-Based {{Control}} of {{Euler-Lagrange Systems}}},
  author = {Ortega, Romeo and Lor{\'i}a, Antonio and Nicklasson, Per Johan and {Sira-Ram{\'i}rez}, Hebertt},
  year = {1998},
  month = nov,
  journal = {Journal of Materials Processing Technology},
  volume = {1},
  publisher = {Springer London},
  address = {London},
  issn = {09240136},
  doi = {10.1007/978-1-4471-3603-3},
  isbn = {978-1-84996-852-2},
  pmid = {25246403}
}

@article{Ortega2001,
  title = {Putting Energy Back in Control},
  author = {Ortega, Romeo and {van der Schaft}, Arjan J. and Mareels, Iven and Maschke, Bernhard},
  year = {2001},
  journal = {IEEE Control Systems Magazine},
  volume = {21},
  number = {2},
  pages = {18--33},
  issn = {02721708},
  doi = {10.1109/37.915398}
}

@article{Ortega2004,
  title = {Interconnection and {{Damping Assignment Passivity-Based Control}}: {{A Survey}}},
  shorttitle = {Interconnection and {{Damping Assignment Passivity-Based Control}}},
  author = {Ortega, Romeo and {Garc{\'i}a-Canseco}, Elo{\'i}sa},
  year = {2004},
  month = jan,
  journal = {European Journal of Control},
  volume = {10},
  number = {5},
  pages = {432--450},
  issn = {0947-3580},
  doi = {10.3166/ejc.10.432-450},
  urldate = {2024-09-03}
}

@article{Ortega2008,
  title = {Control by {{Interconnection}} and {{Standard Passivity-Based Control}} of {{Port-Hamiltonian Systems}}},
  author = {Ortega, Romeo and {van der Schaft}, Arjan and Castanos, Fernando and Astolfi, Alessandro},
  year = {2008},
  month = dec,
  journal = {IEEE Transactions on Automatic Control},
  volume = {53},
  number = {11},
  pages = {2527--2542},
  issn = {0018-9286},
  doi = {10.1109/TAC.2008.2006930},
  urldate = {2024-06-19},
  copyright = {https://ieeexplore.ieee.org/Xplorehelp/downloads/license-information/IEEE.html},
  langid = {english}
}

@mastersthesis{Pratt1995b,
  title = {Virtual {{Model Control}} of a {{Biped Walking Robot}}},
  author = {Pratt, Jerry E},
  year = {1995},
  month = aug,
  langid = {english},
  school = {Massacchusetts Institute of Technology}
}

@inproceedings{Pratt1998,
  title = {Intuitive Control of a Planar Bipedal Walking Robot},
  booktitle = {Proceedings. 1998 {{IEEE International Conference}} on {{Robotics}} and {{Automation}} ({{Cat}}. {{No}}.{{98CH36146}})},
  author = {Pratt, J. and Pratt, G.},
  year = {1998},
  month = may,
  volume = {3},
  pages = {2014-2021 vol.3},
  issn = {1050-4729},
  doi = {10.1109/ROBOT.1998.680611}
}

@misc{Rackauckas2021,
  title = {Universal {{Differential Equations}} for {{Scientific Machine Learning}}},
  author = {Rackauckas, Christopher and Ma, Yingbo and Martensen, Julius and Warner, Collin and Zubov, Kirill and Supekar, Rohit and Skinner, Dominic and Ramadhan, Ali and Edelman, Alan},
  year = {2021},
  month = nov,
  number = {arXiv:2001.04385},
  eprint = {2001.04385},
  primaryclass = {cs, math, q-bio, stat},
  publisher = {arXiv},
  urldate = {2023-11-22},
  archiveprefix = {arXiv},
  langid = {english}
}

@article{Rackauckas2021a,
  title = {Generalized Physics-Informed Learning through Language-Wide Differentiable Programming},
  author = {Rackauckas, C. and Edelman, A. and Fischer, K. and Innes, M. and Saba, E. and Shah, V. B. and Tebbutt, W.},
  year = {2021},
  month = nov,
  journal = {MIT web domain},
  urldate = {2024-11-05},
  copyright = {Creative Commons Attribution 4.0 International license},
  langid = {english}
}

@article{Raissi2019,
  title = {Physics-Informed Neural Networks: {{A}} Deep Learning Framework for Solving Forward and Inverse Problems Involving Nonlinear Partial Differential Equations},
  shorttitle = {Physics-Informed Neural Networks},
  author = {Raissi, M. and Perdikaris, P. and Karniadakis, G. E.},
  year = {2019},
  month = feb,
  journal = {Journal of Computational Physics},
  volume = {378},
  pages = {686--707},
  issn = {0021-9991},
  doi = {10.1016/j.jcp.2018.10.045},
  urldate = {2023-08-08},
  langid = {english}
}

@article{Sadeghian2019,
  title = {Constrained {{Kinematic Control}} in {{Minimally Invasive Robotic Surgery Subject}} to {{Remote Center}} of {{Motion Constraint}}},
  author = {Sadeghian, Hamid and Zokaei, Fatemeh and Hadian Jazi, Shahram},
  year = {2019},
  month = sep,
  journal = {Journal of Intelligent \& Robotic Systems},
  volume = {95},
  number = {3},
  pages = {901--913},
  issn = {1573-0409},
  doi = {10.1007/s10846-018-0927-0},
  urldate = {2022-04-28},
  langid = {english}
}

@article{Sandoval2018a,
  title = {Generalized {{Framework}} for {{Control}} of {{Redundant Manipulators}} in {{Robot-Assisted Minimally Invasive Surgery}}},
  author = {Sandoval, J. and Vieyres, P. and Poisson, G.},
  year = {2018},
  journal = {Irbm},
  volume = {39},
  number = {3},
  pages = {160--166},
  publisher = {Elsevier Masson SAS},
  issn = {18760988},
  doi = {10.1016/j.irbm.2018.04.001}
}

@misc{Sandoval2022,
  title = {Neural {{ODEs}} as {{Feedback Policies}} for {{Nonlinear Optimal Control}}},
  author = {Sandoval, Ilya Orson and Petsagkourakis, Panagiotis and {del Rio-Chanona}, Ehecatl Antonio},
  year = {2022},
  month = nov,
  number = {arXiv:2210.11245},
  eprint = {2210.11245},
  primaryclass = {cs, eess, math},
  publisher = {arXiv},
  doi = {10.48550/arXiv.2210.11245},
  urldate = {2023-07-18},
  archiveprefix = {arXiv}
}

@book{Secchi2007,
  title = {Control of Interactive Robotic Interfaces: A Port-{{Hamiltonian}} Approach},
  shorttitle = {Control of Interactive Robotic Interfaces},
  author = {Secchi, Cristian and Stramigioli, Stefano and Fantuzzi, Cesare},
  year = {2007},
  series = {Springer Tracts in Advanced Robotics},
  number = {volume 29},
  publisher = {Springer},
  address = {Berlin},
  isbn = {978-3-540-49712-7},
  langid = {english},
  lccn = {629.892}
}

@book{Sepulchre1997,
  title = {Constructive Nonlinear Control},
  author = {Sepulchre, Rodolphe and Jankovi{\'c}, Mrdjan and Kokotovi{\'c}, Petar V.},
  year = {1997},
  series = {Communications and Control Engineering Series},
  publisher = {Springer},
  address = {London Berlin Heidelberg},
  isbn = {978-3-540-76127-3 978-1-4471-1245-7},
  langid = {english}
}

@book{Siciliano1999,
  title = {Robot {{Force Control}}},
  author = {Siciliano, Bruno and Villani, Luigi},
  year = {1999},
  publisher = {Springer US},
  address = {Boston, MA},
  isbn = {978-1-4615-4431-9},
  langid = {english}
}

@article{Slotine1987,
  title = {On the {{Adaptive Control}} of {{Robot Manipulators}}},
  author = {Slotine, Jean-Jacques E. and Li, Weiping},
  year = {1987},
  month = sep,
  journal = {The International Journal of Robotics Research},
  volume = {6},
  number = {3},
  pages = {49--59},
  publisher = {SAGE Publications Ltd STM},
  issn = {0278-3649},
  doi = {10.1177/027836498700600303},
  urldate = {2022-02-07},
  langid = {english}
}

@article{Smith2002,
  title = {Synthesis of Mechanical Networks: The Inerter},
  shorttitle = {Synthesis of Mechanical Networks},
  author = {Smith, M.C.},
  year = {2002},
  month = oct,
  journal = {IEEE Transactions on Automatic Control},
  volume = {47},
  number = {10},
  pages = {1648--1662},
  issn = {1558-2523},
  doi = {10.1109/TAC.2002.803532}
}

@article{Smith2020,
  title = {The {{Inerter}}: {{A Retrospective}}},
  author = {Smith, Malcolm C.},
  year = {2020},
  month = may,
  journal = {Annual Review of Control, Robotics, and Autonomous Systems},
  volume = {3},
  number = {1},
  pages = {361--391},
  issn = {2573-5144},
  doi = {10.1146/annurev-control-053018-023917}
}

@article{Song2019,
  title = {A {{Tutorial Survey}} and {{Comparison}} of {{Impedance Control}} on {{Robotic Manipulation}}},
  author = {Song, Peng and Yu, Yueqing and Zhang, Xuping},
  year = {2019},
  journal = {Robotica},
  volume = {37},
  number = {5},
  pages = {801--836},
  issn = {14698668},
  doi = {10.1017/S0263574718001339}
}

@book{Spong1989,
  title = {Robot {{Dynamics}} and {{Control}}},
  author = {Spong, Mark W.},
  year = {1989},
  edition = {1st},
  publisher = {John Wiley \& Sons, Inc.},
  address = {USA},
  isbn = {978-0-471-61243-8}
}

@article{Su2019,
  title = {Improved Human-Robot Collaborative Control of Redundant Robot for Teleoperated Minimally Invasive Surgery},
  author = {Su, Hang and Yang, Chenguang and Ferrigno, Giancarlo and De Momi, Elena},
  year = {2019},
  journal = {IEEE Robotics and Automation Letters},
  volume = {4},
  number = {2},
  pages = {1447--1453},
  publisher = {IEEE},
  issn = {23773766},
  doi = {10.1109/LRA.2019.2897145},
  isbn = {9781538630815}
}

@article{Su2020,
  title = {Bilateral {{Teleoperation Control}} of a {{Redundant Manipulator}} with an {{RCM Kinematic Constraint}}},
  author = {Su, Hang and Schmirander, Yunus and Li, Zhijun and Zhou, Xuanyi and Ferrigno, Giancarlo and De Momi, Elena},
  year = {2020},
  journal = {Proceedings - IEEE International Conference on Robotics and Automation},
  pages = {4477--4482},
  issn = {10504729},
  doi = {10.1109/ICRA40945.2020.9197267},
  isbn = {9781728173955}
}

@article{Su2020a,
  title = {Hierarchical {{Task Impedance Control}} of a {{Serial Manipulator}} for {{Minimally Invasive Surgery}}},
  author = {Su, Hang and Yang, Chenguang and Li, Jiehao and Jiang, Yiming and Ferrigno, Giancarlo and Momi, Elena De},
  year = {2020},
  journal = {Proceedings of the 2020 IEEE International Conference on Human-Machine Systems, ICHMS 2020},
  doi = {10.1109/ICHMS49158.2020.9209500},
  isbn = {9781728158716}
}

@article{Takegaki1981,
  title = {A {{New Feedback Method}} for {{Dynamic Control}} of {{Manipulators}}},
  author = {Takegaki, Morikazu and Arimoto, Suguru},
  year = {1981},
  month = jun,
  journal = {Journal of Dynamic Systems, Measurement, and Control},
  volume = {103},
  number = {2},
  pages = {119--125},
  issn = {0022-0434},
  doi = {10.1115/1.3139651}
}

@book{vanderSchaft2017,
  title = {L2-{{Gain}} and {{Passivity Techniques}} in {{Nonlinear Control}}},
  author = {{van der Schaft}, Arjan},
  year = {2017},
  publisher = {Springer International Publishing},
  address = {Cham},
  doi = {10.1007/978-3-319-49992-5},
  isbn = {978-3-319-49991-8}
}

@article{vanderSchaft2020,
  title = {Port-{{Hamiltonian Modeling}} for {{Control}}},
  author = {{van der Schaft}, Arjan},
  year = {2020},
  journal = {Annual Review of Control, Robotics, and Autonomous Systems},
  volume = {3},
  number = {1},
  pages = {393--416},
  issn = {2573-5144},
  doi = {10.1146/annurev-control-081219-092250}
}

@misc{Wotte2022,
  title = {Discovering {{Efficient Periodic Behaviours}} in {{Mechanical Systems}} via {{Neural Approximators}}},
  author = {Wotte, Yannik and Dummer, Sven and Botteghi, Nicol{\`o} and Brune, Christoph and Stramigioli, Stefano and Califano, Federico},
  year = {2022},
  month = dec,
  number = {arXiv:2212.14253},
  eprint = {2212.14253},
  primaryclass = {cs, math},
  publisher = {arXiv},
  urldate = {2023-06-05},
  archiveprefix = {arXiv},
  langid = {english}
}

@book{Zhou1996,
  title = {Robust and Optimal Control},
  author = {Zhou, Kemin and Doyle, John Comstock and Glover, K.},
  year = {1996},
  publisher = {Prentice Hall},
  address = {Upper Saddle River, N.J},
  isbn = {978-0-13-456567-5},
  lccn = {QA402.3 .Z48 1996}
}

@book{Zhou1998,
  title = {Essentials of Robust Control},
  author = {Zhou, Kemin and Doyle, John Comstock and Doyle, John C.},
  year = {1998},
  series = {Prentice {{Hall}} International Editions},
  publisher = {Prentice Hall},
  address = {Upper Saddle River, NJ},
  isbn = {978-0-13-525833-0 978-0-13-790874-5},
  langid = {english}
}

@inproceedings{aghakhani_task_2013,
    title = {Task control with remote center of motion constraint for minimally invasive robotic surgery},
    url = {https://ieeexplore.ieee.org/document/6631412/?arnumber=6631412},
    doi = {10.1109/ICRA.2013.6631412},
    urldate = {2025-02-11},
    booktitle = {2013 {IEEE} {International} {Conference} on {Robotics} and {Automation}},
    author = {Aghakhani, Nastaran and Geravand, Milad and Shahriari, Navid and Vendittelli, Marilena and Oriolo, Giuseppe},
    month = may,
    year = {2013},
    note = {ISSN: 1050-4729},
    pages = {5807--5812},
}

% \vspace{-33pt}
\begin{IEEEbiography}[{\includegraphics[width=1in, height=1.25in, clip, keepaspectratio]{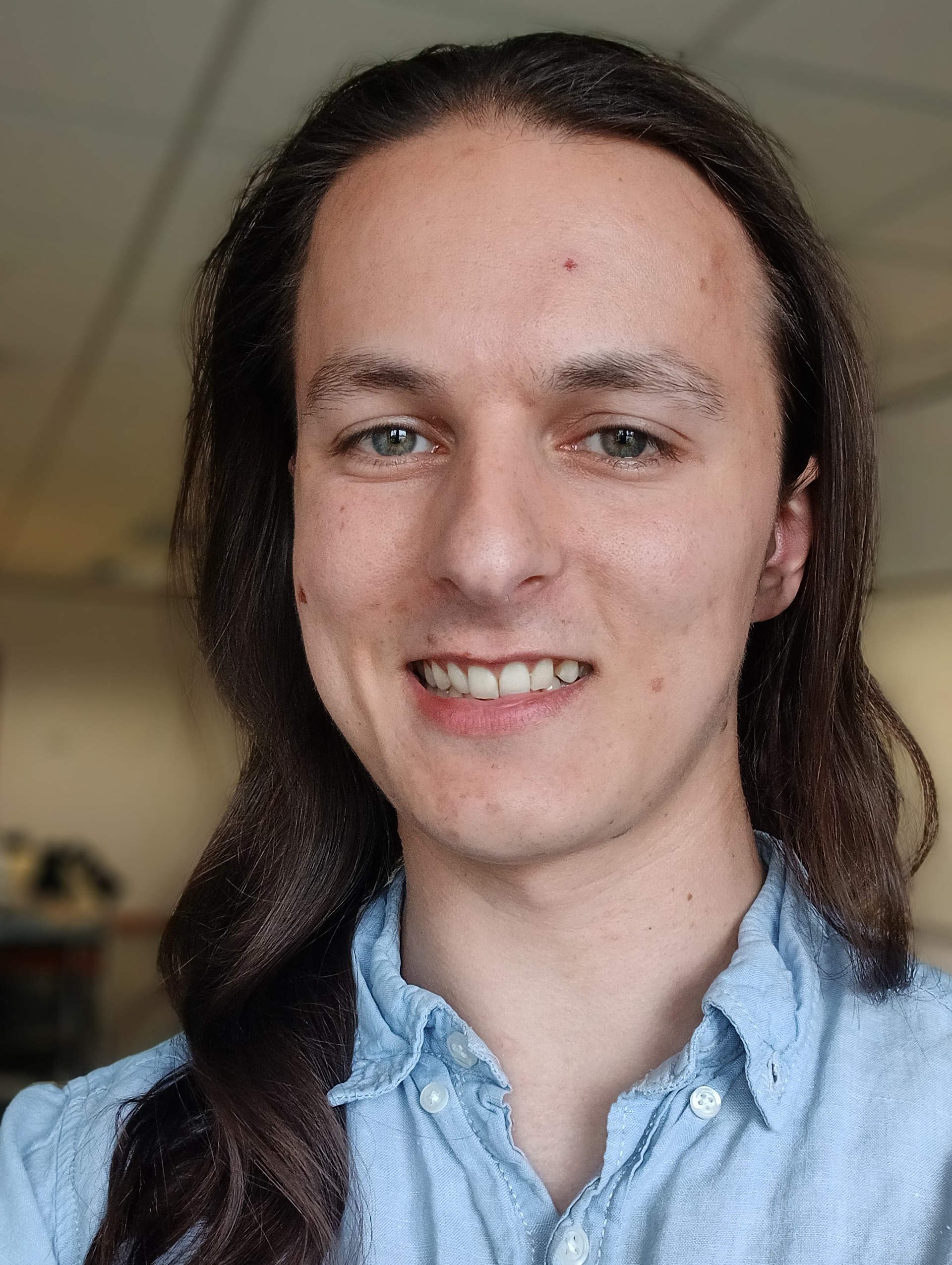}}]{Daniel Larby}
received the MEng degree in control and information engineering in 2020 from the University of Cambridge, UK. He completed his Ph.D. degree in 2025, focussing on robotic control with applications in robotic surgery with the control lab, also in the University of Cambridge, UK.
He now work for Swan EndoSurgical, part of a team working to develop robotic endoluminal surgery solutions.

His research interests include robotics, virtual-mechanism control, passivity based control, impedance control, robotic surgery, and algorithmic differentiation.
\end{IEEEbiography}

\vskip -2\baselineskip plus -1fil
% \vspace{-33pt}
\begin{IEEEbiography}[{\includegraphics[width=1in, height=1.25in, clip, keepaspectratio]{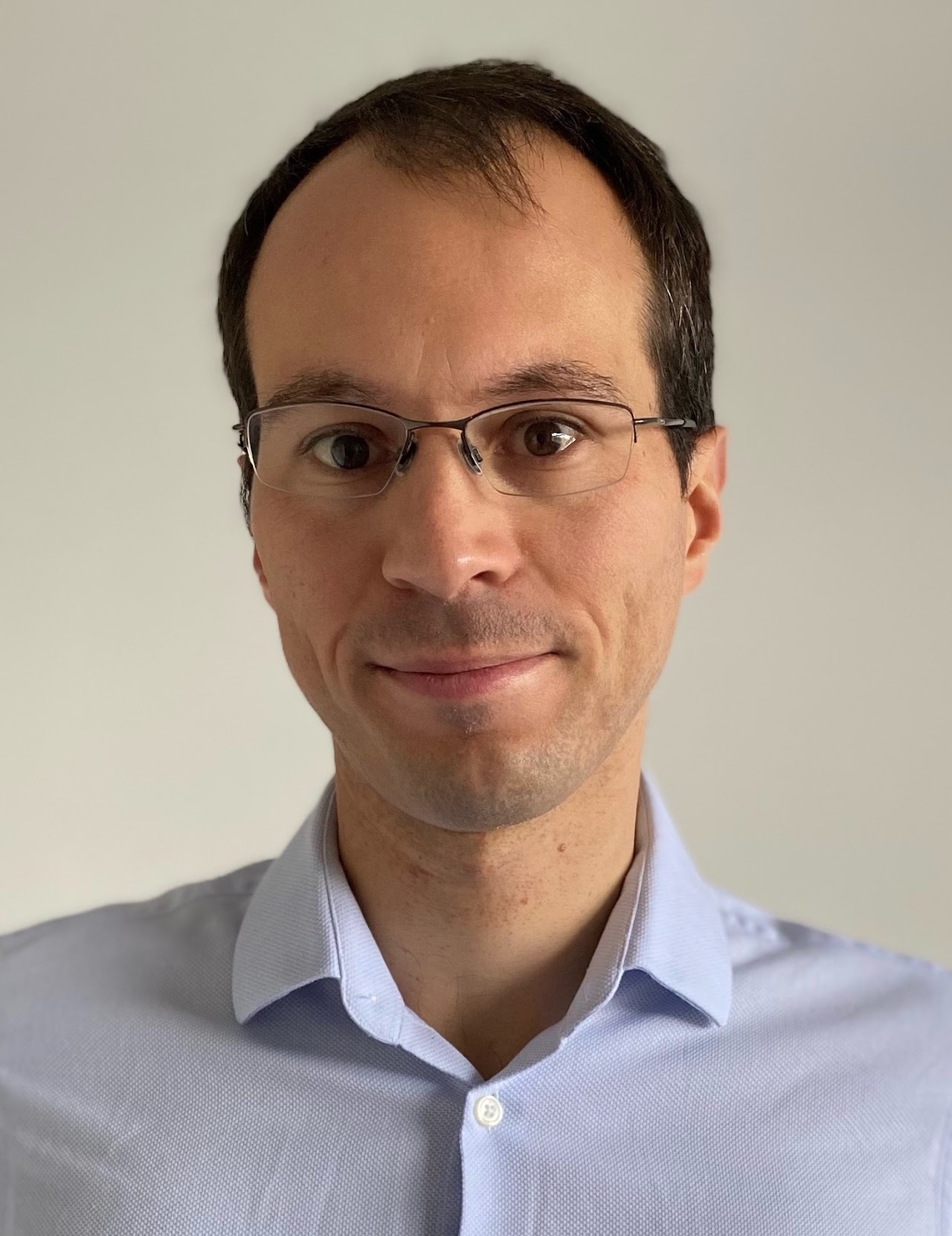}}]{Fulvio Forni}
Fulvio Forni is a Professor of Control Engineering at the University of Cambridge, where he has been a faculty member since October 2015. He earned his PhD from the University of Rome ‘Tor Vergata’ in 2010. Following his doctorate, he conducted postdoctoral research at the University of Liège in Belgium. Forni’s research interests encompass feedback control and robotics. He received the prestigious IEEE CSS George S. Axelby Outstanding Paper Award in 2020. He is a Director of Studies of Newnham College at Cambridge and serves as a co-investigator for the EPSRC Centre for Doctoral Training in Agrifood Robotics ‘Agriforwards’.
\end{IEEEbiography}

\end{document}